\documentclass[journal]{IEEEtran}
\usepackage[utf8]{inputenc}
\usepackage[fleqn]{amsmath}
\usepackage{graphicx}
\usepackage[ruled]{algorithm}
\usepackage[noend]{algpseudocode}
\usepackage{latexsym}
\usepackage{cite}
\usepackage{color,soul}
\usepackage{xcolor}
\ifCLASSINFOpdf
  \DeclareGraphicsExtensions{.pdf,.jpeg,.png}
\else
\fi
\ifCLASSOPTIONcompsoc
 \usepackage[caption=false,font=normalsize,labelfont=sf,textfont=sf]{subfig}
\else
  \usepackage[caption=false,font=footnotesize]{subfig}
\fi

\begin{document}
%
\title{ASP: Learning to Forget with \underline{A}daptive \underline{S}ynaptic \underline{P}lasticity in Spiking Neural Networks}
%
%
%

\author{Priyadarshini~Panda, \textit{ Student Member}, \textit{IEEE}, Jason~M.~Allred, \textit{ Student Member}, \textit{IEEE}, Shriram~Ramanathan,
        and Kaushik~Roy,~\IEEEmembership{Fellow,~IEEE}
\thanks{P. Panda, J. Allred, and K. Roy are with the School
of Electrical and Computer Engineering, Purdue University, West Lafayette,
IN, 47907 USA. S. Ramanathan is with the School of Materials Engineering, Purdue University, West Lafayette,
IN, 47906. Correspondance E-mail: pandap@purdue.edu}
}

\maketitle

\begin{abstract}
A fundamental feature of learning in animals is the \textit{``ability to forget''} that allows an organism to perceive, model and make decisions from disparate streams of information and adapt to changing environments. Against this backdrop, we present a novel unsupervised learning mechanism ASP (Adaptive Synaptic Plasticity) for improved recognition with Spiking Neural Networks (SNNs) for real time on-line learning in a dynamic environment. We incorporate an adaptive weight decay mechanism with the traditional Spike Timing Dependent Plasticity (STDP) learning to model adaptivity in SNNs. The leak rate of the synaptic weights is modulated based on the temporal correlation between the spiking patterns of the pre- and post-synaptic neurons. This mechanism helps in gradual forgetting of insignificant data while retaining significant, yet old, information. ASP, thus, maintains a balance between forgetting and immediate learning to construct a stable-plastic self-adaptive SNN for continuously changing inputs. We demonstrate that the proposed learning methodology addresses catastrophic forgetting while yielding significantly improved accuracy over the conventional STDP learning method for digit recognition applications. Additionally, we observe that the proposed learning model automatically encodes selective attention towards relevant features in the input data while eliminating the influence of background noise (or denoising) further improving the robustness of the ASP learning.  
\end{abstract}

%
\section{Introduction}
With the proliferation of intelligent devices, including smartphones and the Internet of Things, and the resultant explosion of digital data that they generate, computing platforms across the spectrum will increasingly need to acquire, process and analyze new data. This requires such resource-constrained devices to continually extract structures, patterns, and meaning from raw and unstructured datasets in real-time, unsupervised dynamically changing environments.

While advances in deep learning and other machine learning techniques have led to computers matching or surpassing human performance in several tasks (recognition, analytics, inference) \cite{he2015delving, silver2016mastering, lecun2015deep}, the inherent learning mechanisms for such tasks are static. That is, the learning methods use data points from past or old experience to build a predictor (classifier, regression model, recurrent time series model) for processing future behavior. The predictor does not adjust itself (self-correct or adapt) as new events happen. However, accurate prediction based on the new incoming data characteristics with limited memory bandwidth and computational resources needs to be ensured to enable on-chip intelligence. This requires continuous learning over a long period of time with the ability to evolve structural components on demand and forget data becoming obsolete. Based on this, we introduce a novel adaptive self-learning scheme: ASP (Adaptive Synaptic Plasticity) that forgets (or weakens) already learnt connections to make room for new information to adapt to the continuously changing inputs.

Forgetting is essential for learning new things. Recent work in \cite{madronal2016rapid} suggests that the brain is actively erasing memories while learning to continuously process the new input stimuli. At a preliminary level, learning in the brain involves making synaptic connections and re-enforcing them with repeated exposure to a given input stimuli. However, due to limited space available, the brain forgets already learnt connections, gradually, to associate them with new data. In contrast, all computing systems learnt using static mechanisms suffer from ``\textit{catastrophic forgetting}'' where the exposure of an already trained system to new information results in severe loss of previously learned information. It has been shown that this extreme loss of information is not due to the constrained memory capacity (amount of resources available for learning) of the learning system, but is caused by the overlap of representations of information within the system \cite{french1992semi, goodfellow2013empirical, french1999catastrophic}. To address this, our learning scheme maintains a balance between continuous learning and “forgetting” that is necessary to deal with dynamic environments. 

For many years, Artificial Neural Networks (ANNs, including deep learning networks) have dominated across all machine learning models due to their unprecedented performance (in terms of classification/recognition accuracy) in a wide range of computer vision and related applications. However, for applications that require real-time interaction with the environment, the repeated and redundant update of large number of units (neural and weight connections) becomes a bottleneck for efficiency. An alternative has been proposed in the form of Spiking Neural Networks (SNNs), an emergent class of computing paradigm in theoretical neuroscience and neuromorphic engineering \cite{maass1997networks, merolla2014million}. Driven by brain-like asynchronous event based computations, SNNs focus their computational effort on currently active parts of the network, effectively saving power on remaining idle parts, thereby achieving orders of magnitude lesser power consumption in comparison to their ANN counterparts \cite{benjamin2014neurogrid, park201465k}. In addition, SNNs are equipped with self-learning capabilities like Spike Timing Dependent Plasticity (STDP) \cite{song2000competitive} that further make them desirable for adaptive on-chip implementations.

STDP is an associative form of synaptic plasticity that modulates the synaptic strength of a connection based on the temporal correlation between the spiking patterns of pre and post neuron pairs. Several neuroscience theories have demonstrated the role of STDP for Long Term Potentiation (LTP) of synapses that underlie long-lasting memory formation and learning in the brain \cite{martin2000synaptic}. However, an SNN learnt with STDP alone 
does experience catastrophic forgetting while trying to learn quickly in response to a changing environment. Memory persistence is a particularly prominent problem with STDP, as in its naive form STDP implies that any pre/post spike pair can modify the synapse, potentially erasing past memories abruptly.  Grossberg et. al \cite{grossberg1982does, carpenter1988art} have called the problem where the brain learns quickly and stably without catastrophic forgetting its past knowledge while gradually adapting to the flood of new data,  the \textit{stability-plasticity dilemma}. 

The ASP learning proposed in this work addresses the catastrophic forgetting and stability-plasticity dilemma by incorporating a gradual ``decay mechanism'' with the temporal STDP based weight update procedure. In fact, several works on hippocampal learning have demonstrated memory trace decaying over a period of time \cite{hardt2014glua2, villarreal2002nmda, frankland2013hippocampal,madronal2016rapid}. Specifically, LTP that is involved in maintaining memory has been shown to be not permanent and such potentiated responses eventually decay to certain baseline levels \cite{villarreal2002nmda, frankland2013hippocampal}. While the underlying mechanisms for the LTP decay are not known, certain neuroscientists hypothesize that the decay is correlated with ``forgetting'' \cite{hardt2014glua2, villarreal2002nmda, frankland2013hippocampal}. Inspired from this biological evidence, we integrate a weight leaking mechanism with the standard STDP learning rule to model adaptivity in SNNs for pattern recognition applications.  ASP facilitates the gradual degradation or forgetting of already learnt weights to realize new and recent information while preserving some memory or knowledge about the old data that are significant. To the best of our knowledge, the proposed ASP is the first of its kind biologically plausible learning paradigm that effectively combines ``forgetting with learning'' and tremendously boosts the capability of traditional SNNs to deal with dynamic environments. 



Essentially, in an SNN that is learnt using STDP for pattern recognition, 
the weight states are static or are not altered until a post/pre neuron spike is observed. This static learning of weights results in overlap of representations leading to catastrophic forgetting when presented with frequently changing inputs. The key idea of ASP is to leak the weights at every time instant (towards a baseline value) irrespective of post/pre-neuron spikes. The leak rate for each synaptic connection (or leak time constant) is modulated based on the degree of potentiation or depression of a synapse obtained from STDP. While STDP helps in learning the input patterns, the retention of old data/gradual forgetting is attained with  the weight decay. Thus, ASP maintains a balance between forgetting and immediate learning to construct a \textit{stable-plastic} self-adaptive SNN for dynamic environments. 

In addition to dynamic learning, the ``learning to forget'' mechanism achieved with ASP further enables an SNN to perform unsupervised selective attention and denoising by focusing on the task relevant information in the data. In fact, our results (in section 5.4) for handwritten digit recognition from noisy inputs show that the ASP trained SNNs extract distinct digit representations while filtering out the irrelevant background noise. As we will see later, the adaptive weight decay mechanism helps in eliminating the noisy elements in an image as the weights corresponding to those pixels are gradually forgotten. This result emphasizes the significance of the adaptive weight decay for an improved and robust unsupervised learning procedure. 

We would, further, like to note that the initial concept of ASP has been introduced in our recent work that focuses on a quantum perovskite device \cite{zuo2017habituation} that emulates ASP-type characteristics. In fact, the weight decay mechanism in ASP was inspired from the exponential relaxation in conductance observed in the perovskite. However,  the focus of our earlier work was towards the device functionality and the plausibility of the quantum phenomenon to emulate synaptic characteristics. In contrast, this work focuses on the learning rule (or algorithm specifically), how we tackle the key issues in conventional learning models and give more descriptive results and intuitions towards the formulation of this novel learning. Hence, any overlapping discussion with \cite{zuo2017habituation} is primarily meant to introduce the relevant material and mathematical formalism in this manuscript for the sake of completeness and clarity. Essentially, this work (with its algorithmic focus) is aimed at demonstrating the effectiveness of our novel ASP rule as a general unsupervised learning model for future neuromorphic applications. 

\begin{figure}[!t]
\centering
\subfloat[]{\includegraphics[width =0.25\textwidth]{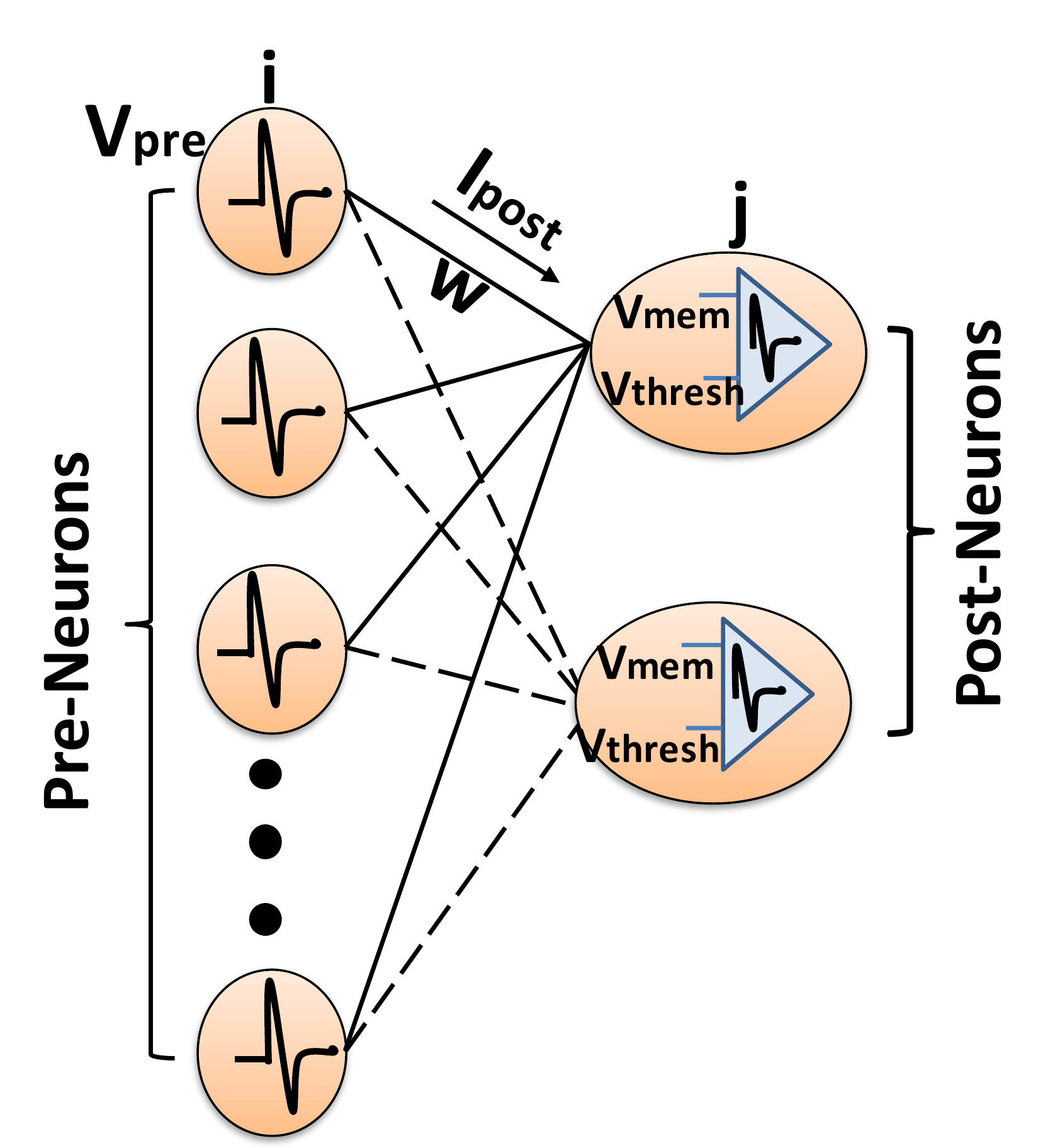}
\label{(a)}}
\hfill
\subfloat[]{\includegraphics[width =0.5\textwidth]{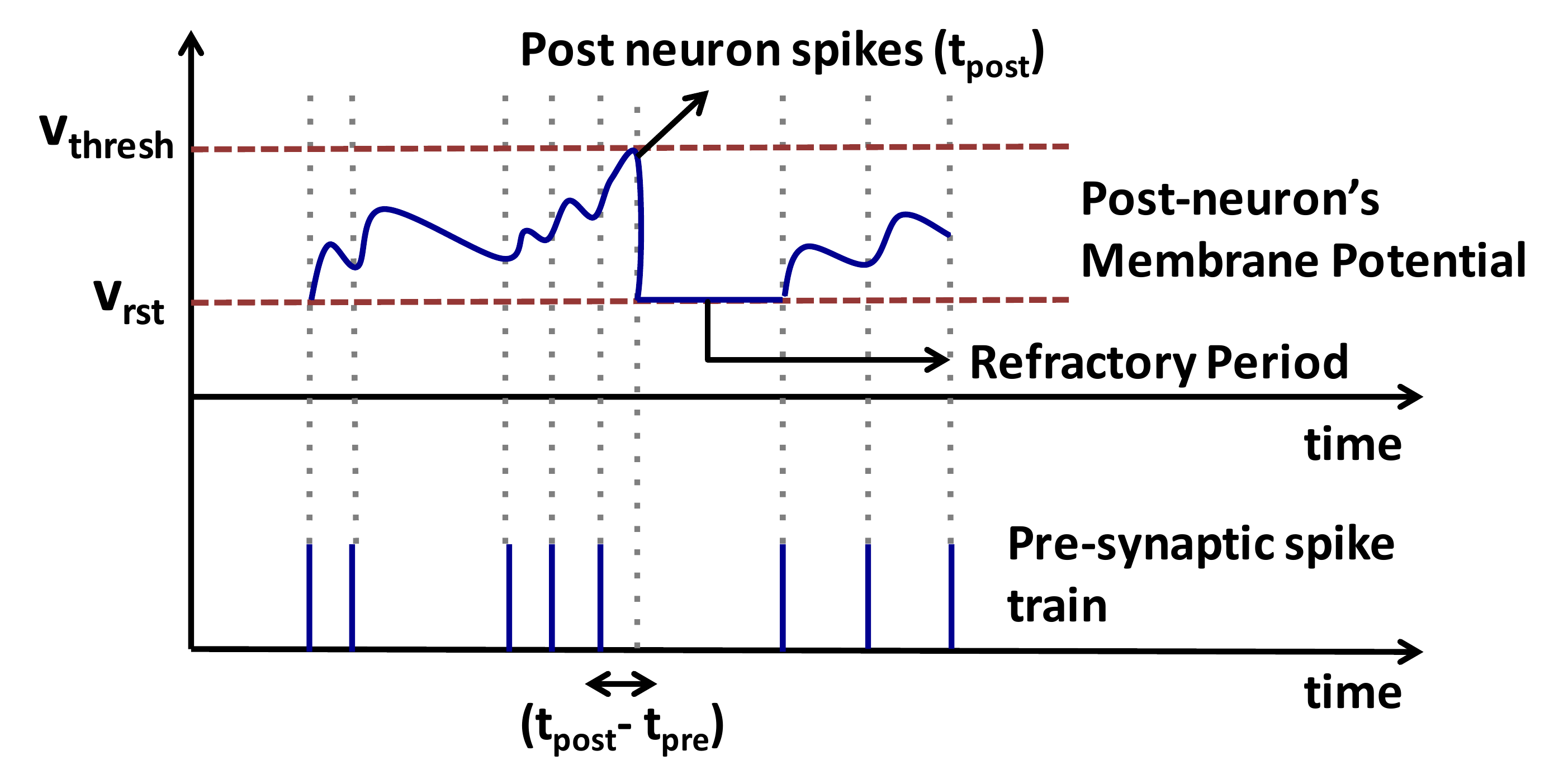}
\label{(b)}}
\caption{(a) A typical SNN architecture consisting of pre-neurons and post-neurons interconnected by synapses. The pre-synaptic voltage spike $V_{pre}$ is modulated by the synaptic weight, $w$, to get the resultant post-synaptic current, $I_{post}$. The post-neuron integrates the current from each interconnected pre-neuron that causes its membrane potential, $V_{mem}$, to increase and spikes when the potential crosses a certain threshold, $V_{thresh}$. (b) The Leaky-Integrate-and-Fire dynamics of the membrane potential of a post-neuron that increases upon the arrival of pre-synaptic spike and decays subsequently. The post-neuron fires when the potential exceeds the threshold $V_{thresh}$. the potential is then reset to $V_{rst}$ and a refractory period ensues during which the neuron is prohibited from firing. The relative timing of the post-neuron and pre-neuron spikes ($t_{post}-t_{pre}$) determines the synaptic potentiation.}
\end{figure}

\section{Spiking Neural Network: Fundamentals}
\subsection{Spiking neuron and synapse model}
The SNN topology used in this work for pattern recognition consists of a couple of layers of spiking neurons interconnected by synapses as standard fully connected ANNs shown in Fig. 1 (a). However, unlike conventional ANNs where a vector is given at the input layer once and the corresponding output is produced after processing through several layers of the network, SNNs require the input to be encoded as a stream of spike events. The neurons transmitting the spikes are referred to as pre-neurons. Each pre-neuronal spike is modulated by the synaptic conductance to produce a resultant post-synaptic current that is received by the post-neurons. 

\begin{equation}
\tau_{post}\frac{dI_{post_{j,i}}}{dt} = -I_{post_{j,i}} + w_{j,i} E_{pre_i}
\end{equation}

where $I_{post_{j,i}}$ is the current received by the jth post-neuron due to a spike event $E_{pre_i}$ at the $i^{th}$ pre-neuron, that are interconnected by a synapse of strength $w_{j,i}$. The post-synaptic current increases by the synaptic weight $w_{j,i}$ instantaneously when a pre-neuronal spike occurs at $t_{pre}$, and subsequently decays with time constant $\tau_{post}$. The spiking neuron model used in this work is the Leaky-Integrate-and-Fire (LIF) model as illustrated in Fig. 1 (b). A simplistic LIF model can be described as follows:

\begin{equation}
\tau_{mem}\frac{dV_{mem_{j}}}{dt} = -V_{mem_{j}} + \sum_i I_{post_{j,i}}
\end{equation}

where $V_{mem_j}$ is the membrane potential of the $j^{th}$ post-neuron. The post-neuron sums the total post-synaptic current leading to an increase in its membrane potential that eventually leaks exponentially with time constant $\tau_{mem}$. It fires or emits a spike when its membrane potential reaches a specific threshold, similar to biological neurons, thus adding an asynchronous component of computation to SNNs. The time instant of occurrence of post-neuron spike is denoted by $t_{post}$. After each post firing event, the neuron's membrane potential is reset and a refractory period ensues during which the neuron is incapable of firing even if additional input spikes are received. 

\subsection{SNN topology for Pattern Recognition}
\begin{figure}[!t]
\centering
\includegraphics[width = 0.4\textwidth]{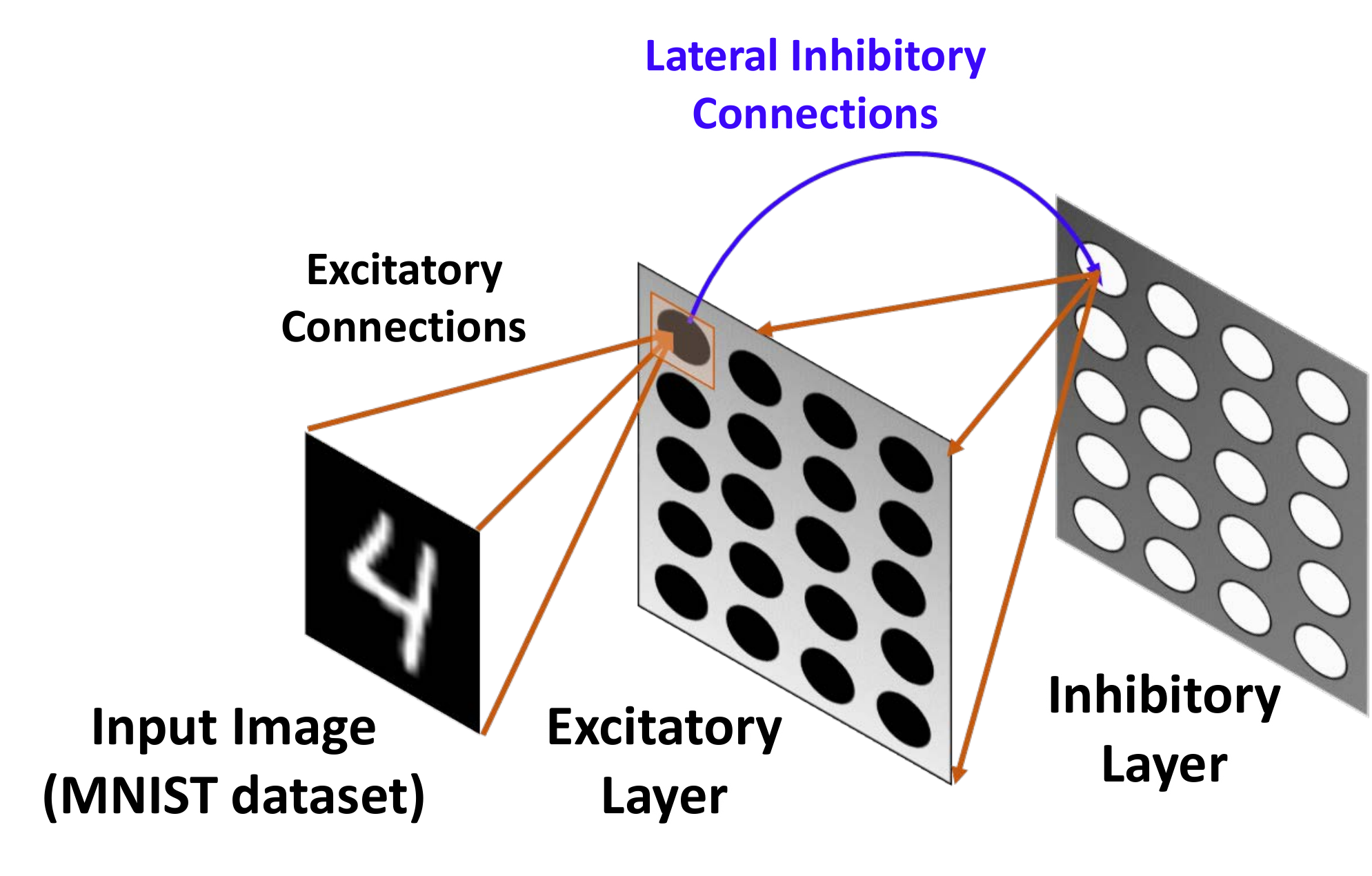}
\caption{SNN topology for pattern recognition consisting of input , excitatory and inhibitory layers arranged in a hierarchical fashion. The input layer is fully connected to the excitatory neurons, that are connected to the corresponding inhibitory neurons in a one-to-one manner. Each of these neurons inhibits the excitatory layer neurons except the one from which it receives the one-to-one connection.}
\end{figure}
The experimental tests in this paper are digit recognition tasks conducted with the MNIST dataset \cite{lecun1998gradient}. A hierarchical SNN structure as shown in Fig. 2 is used. The architecture consists of an input layer followed by excitatory and inhibitory layers. The input layer (28x28 dimension) constitutes the pixel image data for the various patterns in the dataset. Each pixel value is converted to Poisson-distributed spike trains with rates proportional to corresponding pixel intensities. The input layer is fully connected to the neurons in the excitatory layer. Thus, each excitatory neuron has 784 weighted synaptic connections from the input, that are trained (or learnt) to classify a specific input pattern or digit. The excitatory layer is connected in a one-to-one fashion to the inhibitory layer neurons i.e. a spike event in an excitatory neuron would cause the corresponding inhibitory neuron to fire.  Every inhibitory neuron is connected back to all excitatory neurons, except the one from which it receives a forward connection. This connectivity structure provides lateral inhibition that limits the simultaneous firing of various excitatory neurons in an unsupervised learning environment and promotes competitive learning. 

Besides lateral inhibition, we employ an adaptive membrane threshold mechanism called homeostasis \cite{zhang2003other} that regulates the firing threshold to prevent a neuron from being hyperactive. Specifically, each excitatory neuron’s membrane threshold is not only determined by $v_{thresh}$ but by $v_{thresh}+\theta$, where $\theta$ is increased each time the neuron fires and then decays exponentially \cite{querlioz2013immunity}. If the neuron is too active in a short time window, the threshold grows gradually and in turn, the neuron requires more input to spike in the near future. Homeostasis, thus, equalizes the firing rate of all neurons preventing single neurons from dominating the response. Note, the SNN topology shown in Fig. 2 is used to conduct all recognition experiments detailed in the later sections of the paper. 

Next, we discuss the standard STDP model for unsupervised on-line learning in SNNs and then, describe our proposed decay based Adaptive Synaptic Plasticity learning.

\section{Existing Learning Model for SNN}
\subsection{Spike Timing dependent Plasticity (STDP) and its limitations}
\begin{figure}[!t]
\centering
\includegraphics[width = 0.3\textwidth]{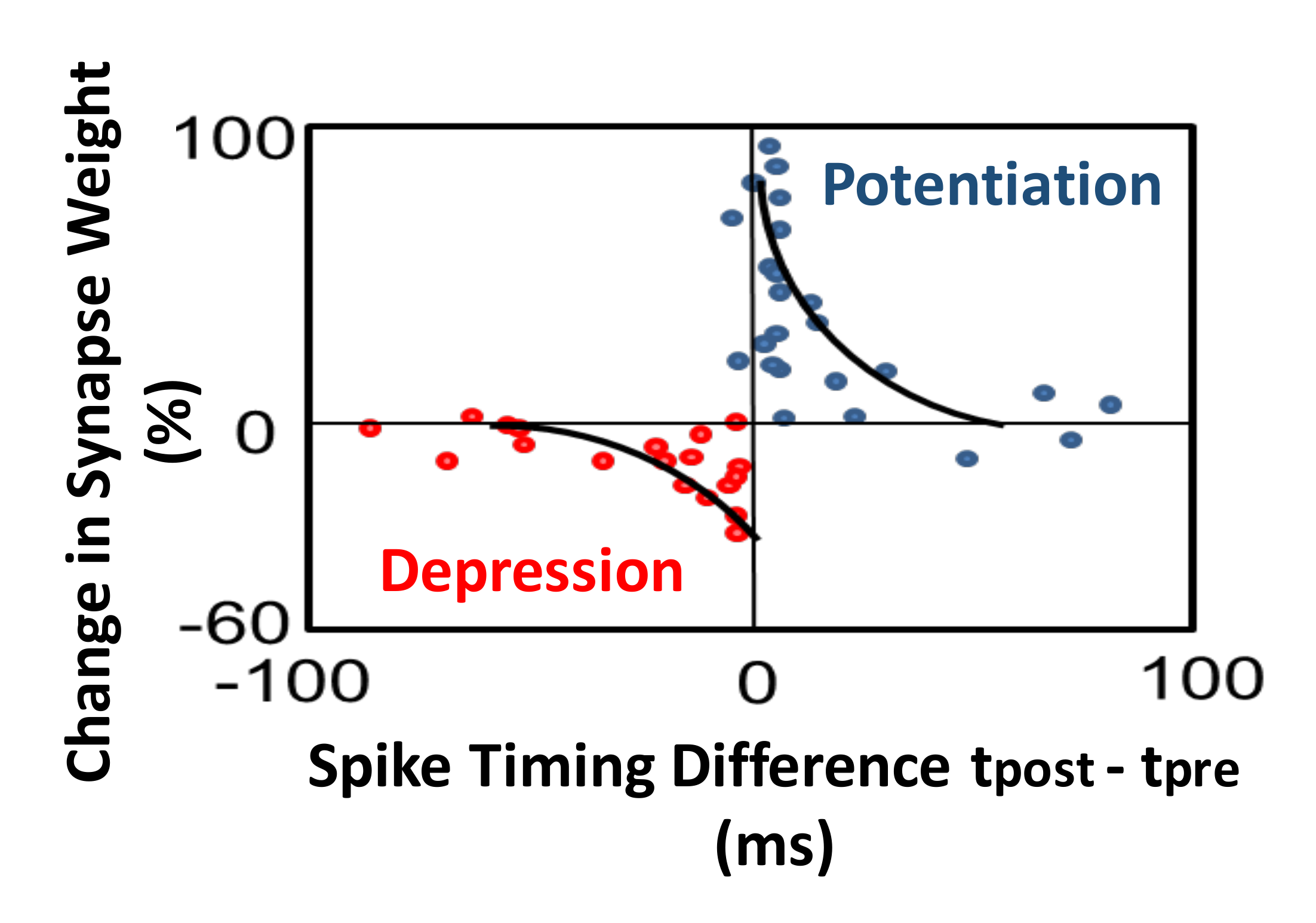}
\caption{Spike timing dependent plasticity curve showing the original measurement from biology in rat's hippocampus \cite{bi1998synaptic} and the traditional model. The relative change in synaptic weight is exponentially related to the difference in spike times of a post-neuron ($t_{post}$) and pre-neuron ($t_{pre}$).}
\end{figure}
In the SNN topology described above, the synapses connecting the input pre-neurons to the post neurons in the excitatory layer need to be trained to learn a generic representation of a class of input patterns. STDP is a set of Hebbian learning rules with mathematical formulation based on the biological evidence observed in the rat hippocampus \cite{bi1998synaptic} as follows:

\begin{equation}
\begin{split}
\Delta w &\propto exp(\Delta t/\tau^- ) \hspace{6mm} if \Delta t < 0\\
             &\propto exp(-\Delta t/\tau^+) \hspace{4mm} if \Delta t \ge 0 
\end{split}
\end{equation}


where $\Delta t =(t_{post} - t_{pre})$ is the time delay between the post and presynaptic spike, $\tau^{+/-}$ is the time constant for potentiation/depression. There are several variations of STDP learning that have been demonstrated to be suitable for learning spatio-temporal patterns. Fig. 3 shows the traditional exponential STDP model with the potentiation and depression window for weight learning \cite{masquelier2009competitive, song2000competitive, abbott1999temporally} and the original measurement from biology \cite{bi1998synaptic}. Synaptic plasticity (or weight change) depends on the interval of time elapsed between pre- and post-synaptic spikes as indicated in Fig. 1(b). The weight is increased/potentiated if a post-neuron fires subsequently after a pre-neuron, with shorter time intervals resulting in an exponentially larger change. On the other hand, it is decreased/depressed if a post-neuron fires before a pre-neuron indicating the absence of a causal relationship. 

When multiple neurons are organized in a simple competitive SNN topology, the STDP learning will enable the network to learn multiple distinct patterns. 
However, in case of on-line learning in a non-stationary environment, a fixed-size network (fixed in terms of number of neurons in excitatory layer and corresponding synaptic connections) must continually alter its response to process new data. Here, competition would not preclude already learnt neurons from responding to a new pattern and adjusting synaptic weights accordingly. This will often result in overlap of representations unless the network size is increased. But circuit/hardware and power limitations pose resource limitations on network sizes where storage cannot be increased. So, constant exposure to new information would result in cumulative synaptic changes leading to a dramatic loss of stored information (or Catastrophic Forgetting) as the previous changes in synaptic weight will be effectively lost. This effect does not occur in the brain since the information loss is gradual. To retain previously learnt information and avoid overlap or overwriting, a common solution proposed is to provide data reinforcement both during initial training and any subsequent training \cite{muhlbaier2004incremental} as discussed next. 

\subsection{Mitigating Catastrophic forgetting in STDP learnt models with data reinforcement}
Data reinforcement is presenting old information to the network along with the new or current information to ensure that previously learnt representation is stably retained while new information is learnt. This helps in resolving the stability-plasticity dilemma of SNNs in a dynamic input environment. Let us discuss two scenarios for digit recognition: a) In Scenario `A', the digits `0' through `9'  are presented one by one sequentially i.e. first all the images for digit `0' followed by all images for digit `1', and so on till digit `9' and thus can be treated as a dynamic learning environment. Here, periodically after a given number of iterations, the SNN is presented with a new class or digit. b) In Scenario `B', the digits are all presented randomly in an intermixed fashion and the intermixed selection of classes (or digits) are repeated iteratively. Presenting all the digits in an intermixed manner iteratively is basically performing data reinforcement wherein old information is re-presented with new data so that later or new input data do not replace previous patterns. Effectively, from a traditional neural network perspective, in Scenario `B', we are retraining the network with both the new and the old information when the network has to learn a new class. In contrast, Scenario `A' is an example of learning without data reinforcement as all classes are presented disjointedly without any relevance to other classes. 

\begin{figure}[!t]
\centering
\includegraphics[width=0.45\textwidth]{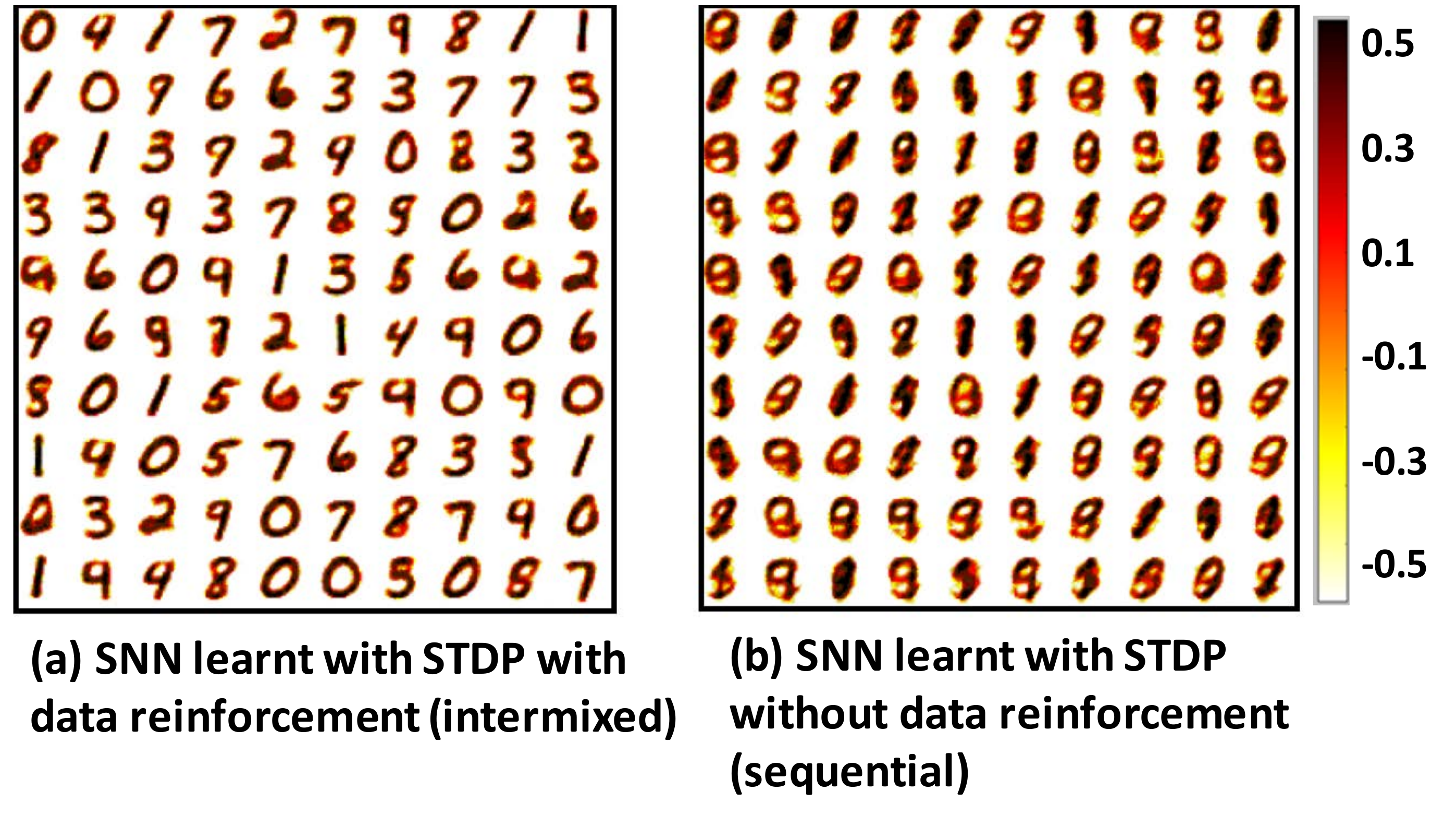}
\caption{Digit representations learnt with STDP in an SNN with 100 excitatory neurons connected to input layer (28x28 pixels) when (a) the digits '0' through '9' are presented in an intermixed manner (i.e. data reinforcement) (b) the digits '0' through '9' are presented sequentially without re-presenting any previous category inputs (i.e. no data reinforcement in dynamic environment). Catastrophic forgetting is observed in SNNs learnt with STDP in a dynamic environment (b) due to significant overlap of representations.}
\end{figure}

Fig. 4 compares an SNN learnt with traditional exponential STDP in the above two scenarios with/without data reinforcement. The SNN topology consists of 100 neurons in the excitatory layer with fully connected synaptic weights from the input (28x28 pixels for MNIST). During learning, the excitatory synaptic weights from the input layer to each excitatory neuron are modulated to learn a particular input digit using the learning rule. Towards the end of the training phase, the weights (or excitatory connections) that are randomly initialized eventually learn to encode a generic representation of the digit patterns. Specifically, the excitatory connection weights (refer to Fig. 2) fanning out of the higher-intesity  (or white)  pixel regions will get potentiated while the weights from the low-intensity regions on the input image will be depressed during the learning phase. Correspondingly, the color-map figures shown in Fig. 4 represent the weight values learnt corresponding to each excitatory neuron when learning stops. 

Fig. 4 (a) clearly shows how the SNN presented with a random mixture of input classes is able to learn a balanced representation of all input patterns due to data reinforcement. Fig. 4 (b) shows the SNN learnt in Scenario ‘A’ where old information or data categories were not re-presented to the network, causing an overlap of input representations. The SNN always tried to learn the new digit representation while trying to retain a portion of the old data. However, fixed network size and absence of data reinforcement resulted in accumulation causing new weight updates to coalesce with already learnt patterns. Hence, an SNN will be rendered useless for categorizing the digits in a dynamic environment without data reinforcement. 
However, in on-line real time learning, it is often impractical and even expensive to store all old data samples for retraining or data reinforced learning each time a new input pattern or class is encountered. ASP, wherein we perform STDP updates augmented with dynamic weight leaking, would enable an SNN (with fixed resources) to gracefully forget already learnt information while retaining the significant patterns and gradually adapt to new data in an online learning environment. 

Next, we discuss the principles of weight decay based learning and how it can be applied for pattern recognition with SNNs.

\section{Adaptive Synaptic Plasticity for Learning to Forget}

\subsection{Adaptivity with Weight Decay in SNNs}
Before we discuss the proposed ASP learning rule, let's see how the weight decay alone helps in adapting a network to continuously changing inputs. Here, we do not incorporate any timing-based associative rules such as STDP with the weight decay process. We only perform an exponential decay and recovery of synaptic weights based on the incoming spikes from the pre-neurons as 
\begin{equation}
\tau_{leak} \frac{dw}{dt}= -\alpha w+ P(t) 
\end{equation}

where $\tau_{leak}$ (\textit{200 ms}) is the time constant for the exponential decay of weights, $\alpha$ (\textit{0.01}) determines the overall learning rate. Here, $P(t)$ denotes the presynaptic trace that models the recent presynaptic history. To improve simulation speed, the weight dynamics are computed using synaptic traces \cite{morrison2007spike, diehl2015unsupervised}. This means that, besides the synaptic weight, each synapse keeps track of the presynaptic trace as well. Each time a pre-neuron fires, the presynaptic trace is increased by 1, otherwise $P(t)$ decays exponentially as 
\begin{equation}
P(t)= exp(-t/\tau_{trace}) + t_{pre}
\end{equation}

where $\tau_{trace}$ (\textit{20 ms}) is the time constant for the decay of pre-synaptic trace and $t_{pre}$ indicates the presence/absence of pre-neuronal spike. The $t_{pre}$ value is 0 if pre-neuronal spike is absent, while 1 when pre-neuronal spike occurs at any time instant $t$. As per Eqn. 4, during a given epoch or time period when an input pattern is presented, the synaptic weights are potentiated proportional to the presynaptic trace value calculated at the particular time instant when a $t_{pre}$ occurs (or pre-neuron fires). At all other time instants, the weights are leaking or decaying. 
Please note that the above learning is unaccompanied by any sort of correlation between post and pre-neuron spike timings. It is solely driven by pre-neuronal spikes. 

\begin{figure}[!t]
\centering
\includegraphics[width=0.45\textwidth]{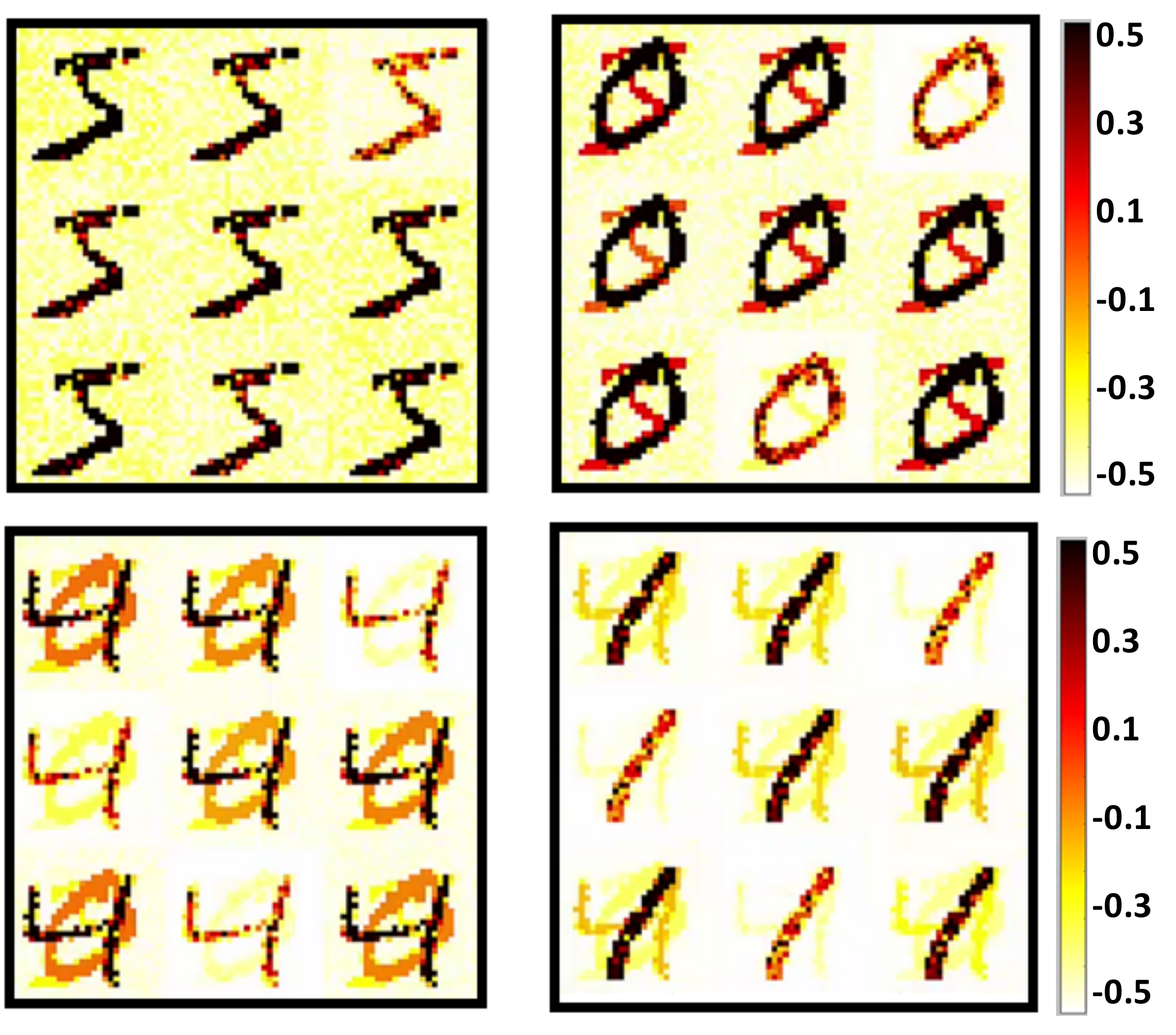}
\caption{Digit representations learnt with isolated weight decay learning in an SNN with 9 excitatory neurons connected to input layer (28x28 pixels). The input digits are presented sequentially (i.e. dynamic environment with no data-reinforcement). The network is overly plastic and adaptive to the continually changing inputs.}
\end{figure}

Fig. 5 shows the digit representations learnt in an SNN's excitatory layer using the isolated decay based weight modulation (Eqn. 4) when different digits (Digit: 5, 0, 4, 1; order chosen randomly) are presented to the network sequentially. The SNN comprises of 9 neurons in the excitatory layer. The color intensity of the patterns are representative of the value of synaptic weights. As different digits are presented, the weights adapt to the new input pattern while forgetting the old information. Thus, we see that the current input representation is learnt on top of the older and most recent pattern. While the weights are highly potentiated for the current input pattern (almost black-to-red intensity), we can observe from the color intensity (orange-to-yellow) that the old information is being forgotten (or corresponding synaptic connections leak due to absence of spike inputs from the pre-neurons in that region as per Eqn. 4). 

A key observation here is that while learning a new pattern, the overlap occurs with the most recent input. For instance, while learning a digit `4', the SNN still retains some information about the previous input digit `0'. However, it has completely forgotten all information regarding the older digit `5'. Thus, weight decay enables the SNN to gracefully forget old information while adapting the network to learn new data. This is what helps the SNN to avoid catastrophic forgetting that was observed with standard STDP learning without data reinforcement (Fig. 4 (b)), wherein the weight updates got accumulated resulting in significant overlap of representations. Thus, it can be inferred that forgetting or weight decay is integral to self-learning and adapting in a dynamic environment. However, in the above learning rule, relative timing of pre- and post-synaptic spikes has not been considered for modifying the weights. The weight updates occur based only on pre-neuronal spike that is in contrast to the commandments of Hebbian learning \cite{morris1999hebb}. Hence, there is no possibility of initiating a competition between neurons to learn different patterns. As a result of this, all the excitatory neurons in Fig. 5 learn uniformly and pick up the same pattern. The SNN in this case is excessively plastic and inadequately stable as it immediately adapts to the new information. Thus, it is essential to incorporate spike timing based correlation with the weight decay to learn in an unsupervised dynamic environment in a \textit{stable-plastic} way. 

Another way of invoking competition is by forcing certain neurons selectively to learn different patterns. However, in that case, we move into the supervised learning domain that is generally undesirable for on-chip implementations as motivated earlier. 

\subsection{Adaptive Synaptic Plasticity: Combining STDP with Weight Decay}
In the weight decay learning discussed above, the role of post synaptic neuron's spiking activity was not accounted for modulating the weights. 
Adaptive Synaptic Plasticity (ASP) blends in associative learning with the adaptation behavior that helps in retention and gradual adaptation to new inputs as well as evokes competition across neurons to learn distinct patterns. Essentially, we modulate the leak time constant (or forgetting behavior) and the exponential recovery of synaptic weights, observed earlier with isolated decay-based learning, using the temporal dynamics of pre- and post-synaptic neurons. In STDP (refer to Fig. 3), the spike timing correlation (or difference of the spike events) between the post/pre neurons determines the window (positive/negative) across which the synaptic modification (potentiation/depression) takes place. Likewise, ASP also incorporates different windows for potentiation and depression based on the firing events of the post/pre neurons as shown in Fig. 6. 

\begin{figure}[!t]
\centering
\includegraphics[width=0.5\textwidth]{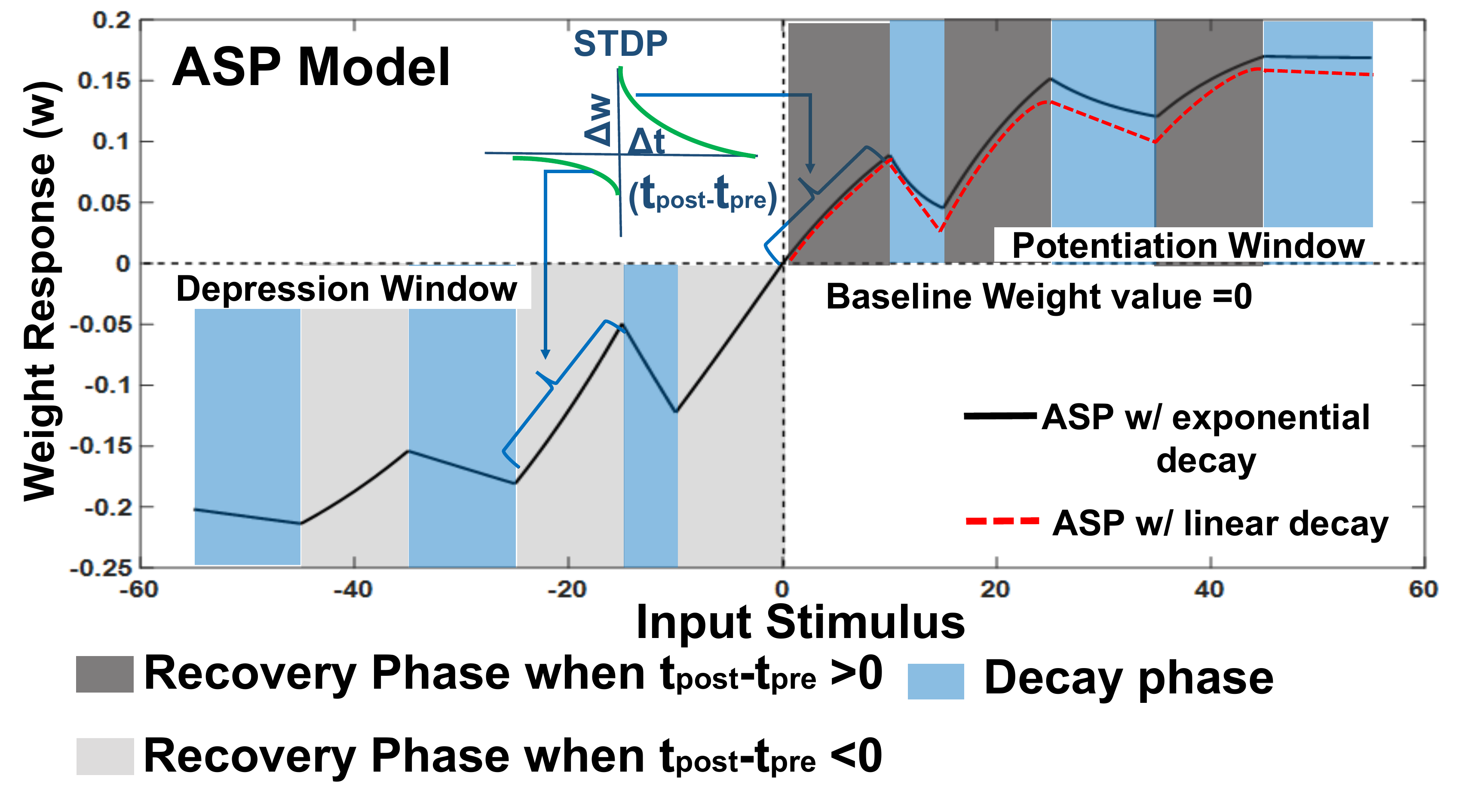}
\caption{ASP model for weight modulation \cite{zuo2017habituation}. During the recovery phase, i.e. when a spiking event occurs at the post/pre-neuron, the weights are potentiated (or depressed) based on the $t_{pre}$ and $t_{post}$ realtive timing difference following the STDP dynamics (Eqn. 7). The weights leak towards baseline value (=0) during the decay phase and the leak time constant is varied based on the post-synaptic neuron's spiking activity (Eqn. 8). Two different ASP models with exponential (black curve) and linear (dotted red curve) decay are shown.}
\label{fig_sim}
\end{figure}

ASP interprets any spiking activity at the post or pre-neuron as the presence of input stimulus to initiate, what we term as, recovery phase for the synaptic weight interconnecting them. The recovery can be an exponential increase (potentiation) or decrease (depression) based on the temporal difference between the spiking activities of the neurons. The potentiation or depression of the weights in the recovery phase is obtained from the exponential STDP formulation (Refer to Eqn. 3) as shown in Fig. 6. Thus, if a pre-neuronal spike forces a post-neuron to fire (i.e. $t_{pre} < t_{post}$), the corresponding synapse should potentiate based on Eqn. 3 since it’s positively correlated with the input pattern. On the contrary, if a pre-neuron fires after a post-neuronal spike (i.e. $t_{pre} > t_{post}$), the specific synapse ought to depress. 

Following the recovery (STDP based weight modulation), the decay or forgetting phase in a synaptic weight ensues, as illustrated in Fig. 6. During the decay phase, the weights leak in both potentiation/depression window towards a baseline value with varying leak time constants. Please note that in ASP, the decay phase (or weight leaking) continues even when there is no spiking activity (or input stimulus) observed at both post/pre neuron. For instance, during the refractory period, when a post neuron does not spike, then in the absence of a pre-neuronal spike the weights will continue to leak. This is in stark contrast to the traditional STDP learning wherein the weight states are only altered in the presence of a pre- or post-neuronal spiking activity. Thus, it can be inferred that the fundamental weight response in ASP is to conduct STDP based weight modifications with intermediary weight decay (or leak) at all time instants during the training period. 

To emphasize the efficacy of ASP for adaptive learning, we validate our experiments with two different kinds of decay: exponential (discussed earlier in Section 4.1) and linear as illustrated by the black and dotted red model, respectively, in Fig. 6. The formulations for each of these decay are explained later. In Fig. 6, the baseline value of weights toward which the weights decay is 0. This implies that the weights are dynamic or continuously changing at every time instant during the training phase when input patterns are presented. Please note, for the sake of convenience in representation and for clarity, the characteristics of ASP with linear decay has not been shown for the depression window in Fig. 6. However, it will follow similar behavior as the other ASP model with weights decaying linearly toward the baseline value.

Fig. 7 shows the synaptic weight modulation using ASP for a given synaptic weight based on the spiking patterns of the interconnecting neurons. 
As in Fig. 6, the baseline value toward which the synapses leak is assumed to be 0 here. Please note that while depression of weights in recovery phase is a result of negative correlation between post and pre-neuron, the leak dynamics in decay phase (only shown for exponential leak) encode the adaptive behavior of the network. The leak dynamics determine which synaptic weights connected to a post neuron (that has learnt old or insignificant data) should be forgotten to learn the new data. On the other hand, the synaptic weight updates performed during the recovery phase potentiation/depression enable an SNN to learn a generic representation of a class of input patterns competetively. For instance, the weights of an excitatory layer post-neuron learning a digit `2' should spike for different instances of `2' so that it learns a more generic representation rather than just mimicking a specific instance. Thus, synaptic depression (based on STDP) and leak (based on weight decay) have different roles in ASP learning.

Note, Fig. 6, 7 are animations that show the weight modification with ASP based on presence/absence of spiking activity. For the sake of clarity, we demarcated the ASP rule into two phases since the weight update rule in the presence of a pre/post spike is different than when spikes are absent. Input stimulus on the X-axis basically indicates the presence (or absence) of pre/post synaptic spike that initiates a recovery (or decay) phase. It quantifies the timing correlation ($t_{post}-t_{pre}$) in a post/pre neuronal pair. 

\begin{figure}[!t]
\centering
\includegraphics[width=0.5\textwidth]{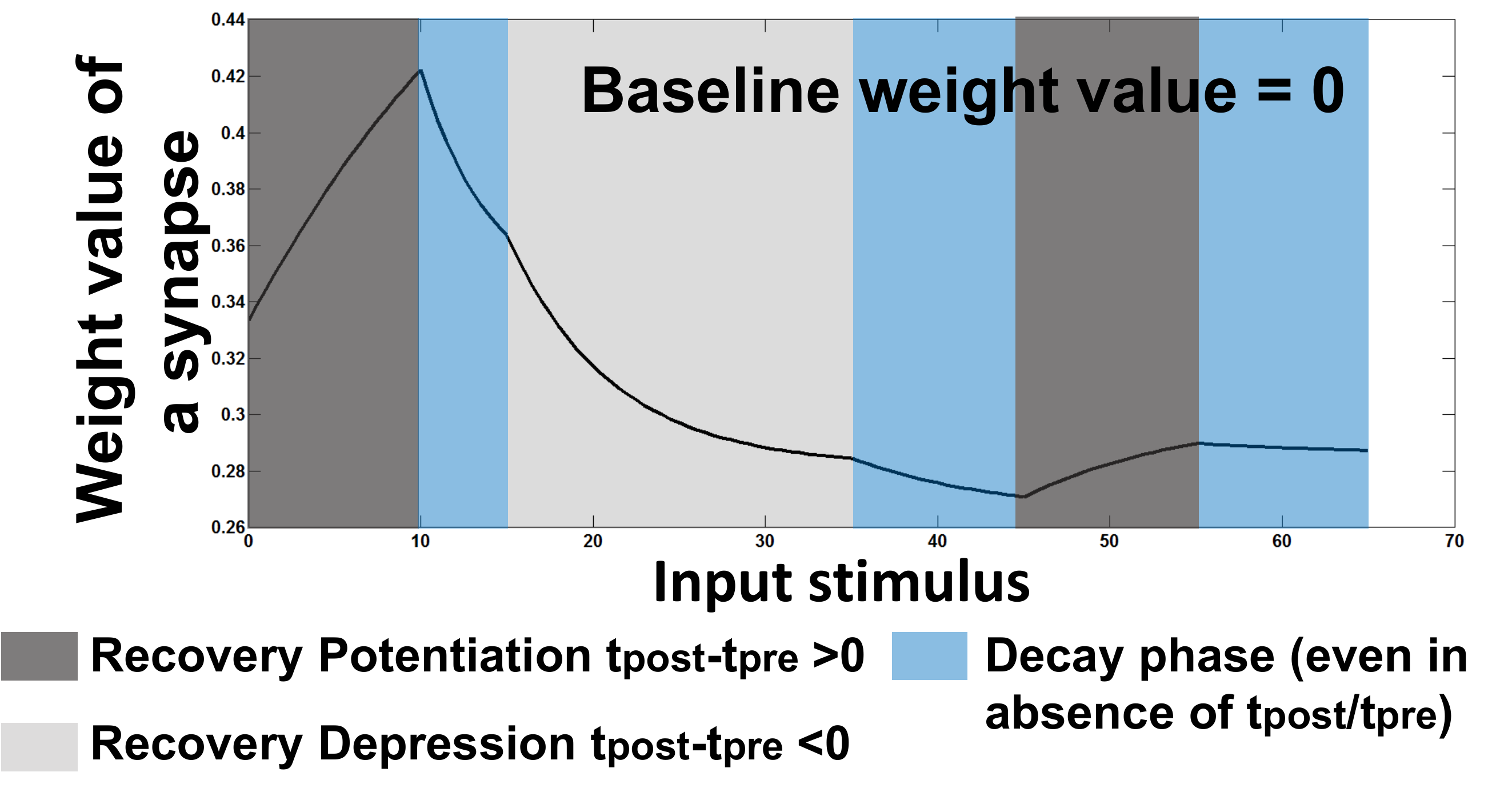}
\caption{Weight response behavior of a particular synapse during intermediate recovery (potentiation or depression) and decay phases.}
\end{figure}

\subsection{ASP: Learning Rules}
Now, we will discuss the mathematical formulations for ASP and understand the temporal dynamics related insights that eventually enables the SNN to learn to forget irrelevant old information while retaining significant data and adapting to new patterns. Note, the formalisms for ASP have already been presented in \cite{zuo2017habituation}. The same mathematical formalisms have been discussed here with additional insights and descriptive logical relevance of the chosen parameters to the dynamic learning scenario.

\subsubsection{Recovery Phase}
As described earlier, the weight dynamics are calculated using synaptic traces. In ASP learning, we maintain three different kinds of traces, for each synaptic weight, to track the spiking activities of the corresponding pre- and post-synaptic neurons:
\begin{itemize}
\item Recent presynaptic trace ($Pre_{rec}$) that does not accumulate over time (only accounts for the most recent spike)
\item Accumulative presynaptic ($Pre_{acc}$) trace that adds over time (accounts for the entire spike history of the presynaptic neuron for a given time period or epoch during which a particular pattern is presented to the SNN)
\item Postsynaptic trace ($Post$) that accumulates over time based on the postsynaptic neuron’s spiking activity. 
 \end{itemize}
The above trace values are computed as shown in Eqn. 6. When any spiking activity is observed the trace value is increased, otherwise it decays exponentially.
\begin{align}
\begin{split}
Pre_{rec}(t) &= exp(\frac{-Pre_{rec}}{\tau_{rec}}) ; \hspace{1mm}  Pre_{rec} =1  \hspace{1mm} at \hspace{1mm}t_{pre}\\
Pre_{acc}(t) &= exp(\frac{-Pre_{acc}}{\tau_{acc}}); \hspace{1mm} Pre_{acc} +=1 \hspace{1mm}at \hspace{1mm} t_{pre}\\
Post(t) &= exp(\frac{-Post}{\tau_{post}}); \hspace{1mm} Post +=1  \hspace{1mm} at \hspace{1mm} t_{post} 
\end{split}
\end{align}

Now, it is evident that in order to account for appropriate accumulative spike history, the time constant for decay of the accumulative pre-trace ($Pre_{acc}$) has to be larger than that of the recent pre-trace ($Pre_{rec}$). In our simulations, $\tau_{acc}=10\tau_{rec}$, $\tau_{post}=2\tau_{acc}$. Now, the weight update during the recovery phase follows the exponential STDP behavior discussed earlier. We use a modified version of the STDP model used in \cite{diehl2015unsupervised} to obtain the weight changes in presence of input stimulus in ASP. When a post-synaptic neuron fires a spike, the weight change $\Delta w$ is calculated based on the presynaptic traces ($Pre_{rec}, Pre_{acc}$) 

\begin{align}
\begin{split}
\Delta w = \eta(t) [(Pre_{rec} - offset) - \frac{k_{const}}{2^{Pre_{acc}}}]\\
\eta(t) = \frac{k_{1,const}}{(Post(t)+1)}
\end{split}
\end{align}

where $\eta(t)$ is a time dependent learning rate inversely proportional to the post-synaptic trace value ($Post(t)$ from Eqn. 6) at a given time instant. The learning rate decreases as the spiking activity of the post-synaptic neuron increases for a given input. This ensures stable and retentive learning of an input pattern by a particular neuron. It also prevents the neuron from quickly adapting to a new pattern (or catastrophic forgetting). The $offset$ ensures that the weights interconnecting those pre-synaptic neurons that rarely lead to firing of a post-synaptic neuron will depress. For instance, in case of digit inputs, the lower intensity pixel regions for a particular digit will become more and more disconnected resulting in lowering of synaptic weight values corresponding to those pre-neurons. In Eqn. 7, the first part represents the potentiation or depression of weights based on the most recent pre-synaptic spike (as the simple STDP model in \cite{diehl2015unsupervised}). However, as seen earlier in Section 3.2, such simplistic weight modification leads to erasure of memory traces. Besides precise spike timings that identify the temporal correlation between input patterns, learning rule should incorporate the significance of the inputs to modulate the weights. 
Input based significance driven learning would enable the SNN to learn in a stable-plastic manner in a dynamic environment.

The second part of Eqn. 7 ($k_{const}/2^{Pre_{acc}}$) quantifies the dependence of the weight change on input significance. We define input significance as directly proportional to the number of training patterns (of a particular input class) shown at the input layer of an SNN. Consequently, for larger number of similar patterns, the corresponding input neurons will have more frequent spikes. In that case, $Pre_{acc}$ value will be high that would eventually make the second term in Eqn. 7 less dominant for determining the final weight update. Thus, for more frequent input spikes at the pre-neuron, the weight update will be more prominent. Hence, the learning rule in the recovery phase encompasses significance of the inputs with synaptic plasticity.

\begin{figure}[t!]
\centering
\includegraphics[scale =0.35]{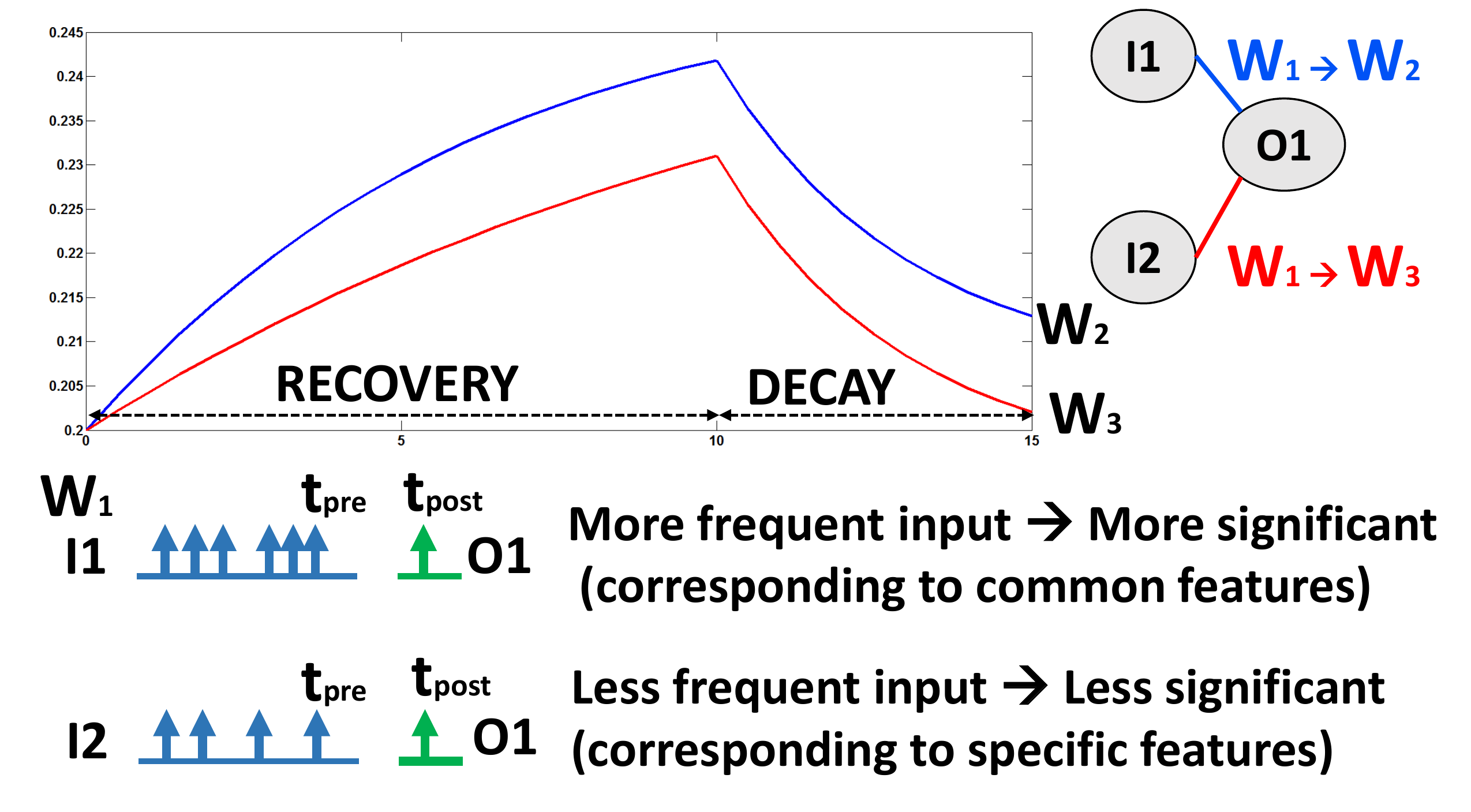}
\caption{Significance driven weight update observed with ASP. More frequent input spikes corresponding to the common features across old and new input patterns will have a greater weight update than the less frequent ones that correspond to specific features for a given input.}
\label{fig8}
\end{figure}

Besides significance driven learning, the second part of Eqn. 7 also enables an SNN to learn more generalized input representations. Fig. \ref{fig8} shows two scenarios wherein the weight updates vary based on the frequency of input spikes. It is seen that the final weight value towards the end of the recovery phase is greater for the frequent input ($I_1$). The prominent weights will essentially encode the features that are common across different classes of old and new inputs as the pre-neurons across those common feature regions in the input image will have frequent firing activity. This eventually helps the SNN to learn more common features with generic representations across different input patterns. We can visualize this as sort of regularization wherein the network tries to generalize over the input rather than overfitting such that the overall accuracy of the network improves. 

Note that during the recovery phase in ASP, the weight updates are triggered only when a spike is fired by a postsynaptic excitatory neuron. Since the firing rate of the post-synaptic neurons is generally low, the weight update does not require many computational resources. Using a learning rule that updates the weights even at pre-synaptic spikes will initiate a weight change for every post neuron in the excitatory layer fully-connected to the pre-neuron. This will be computationally more expensive to simulate in software simulations. 

\subsubsection{Decay Phase}
The decay phase in ASP learning involves weight leak to emulate forgetting of insignificant information to adapt to new data without catastrophic overlap of representations. As discussed earlier, the weights can undergo an exponential/linear decay (refer to Fig. 6) towards a baseline value as
\begin{align}
\begin{split}
\tau_{leak} \frac{dw}{dt}= -\alpha w \hspace{5mm} \textit{Exponential decay}\\
\tau_{leak} \frac{dw}{dt}= -\alpha_{lin} \hspace{5mm}  \textit{Linear decay}\\
\tau_{leak} = k_{2,const}[(Post(t)+1)*2^{(v_{thresh} + \theta)}]
\end{split}
\end{align}
where $\alpha$, $\alpha_{lin}$ are constant decay rates and $\tau_{leak}$ is the time constant of decay. $\tau_{leak}$ is a time dependent quantity that is proportional to the post-synaptic trace value ($Post(t)$ from Eqn. 6), the homeostatic membrane threshold value ($v_{thresh} + \theta$) at a given time instant and the current value of the synaptic weight $w$. 

Earlier, in isolated weight decay learning, we observed that the network was overly plastic modifying its weights for any new pattern. However, in order to retain the learnt information, it is desirable that the corresponding weights should leak less. It is evident that a neuron that has learnt a particular pattern will have a higher synaptic weight. Also, the neuron will exhibit more spiking activity reckoned by the higher post trace value $Post(t)$ that will, subsequently, increase the time constant of decay, $\tau_{leak}$. Higher $\tau_{leak}$ causes the weight to forget less.  The overall leak rate( $\alpha/\tau_{leak}$) decreases with increasing $\tau_{leak}$. 
Here, the value of $Post(t)$ is indicative of how recent and latest the input pattern is (i.e. higher $Post(t)$ observed for most recent input). It does not account for the significance of the input pattern defined in terms of number of times a particular pattern has been presented to an SNN. 

The homeostatic membrane threshold of a post-neuron is representative of the significance of the input pattern. A neuron's membrane threshold ($v_{thresh} + \theta$) will be high only when it is firing more. A neuron spikes more learning a given pattern well, when that pattern is presented several times to the network. Higher membrane threshold qualifies that the post-neuron (in excitatory layer) has learnt a significant input and its corresponponding synaptic connections should leak less. Whilst, the connections to an excitatory neuron that has learnt a pattern that the network has seen fewer number of times are forgotten. The weights corresponding to those neurons should eventually be modified to learn more recent patterns. Hence, the SNN learns to forget insignificant information while trying to learn more recent and retain significant, yet old, data using ASP. 

A key observation here is that the weight leak in the decay phase (irrespective of linear or exponential decay) is dominated by the post-neuron's spiking activity (and membrane threshold). All weights connected to a post-neuron in the excitatory layer will have similar (\textit{not exactly same due to dependence on current weight value}) decay time constant and hence show uniform leak dynamics during the decay phase. On the contrary, during recovery phase, the weight dynamics of each synapse will be different as it is determined by both the post and pre-neuronal spiking activity. In Fig. \ref{fig8}, we see that both the weights connected to the post-neuron $O_1$ exhibit almost same leak. As the weight update in recovery phase was more prominent for $I_1-O_1$ (2nd term in Eqn. 7), it finally settles at a higher value than that of $I_2-O_1$. Thus, the combined recovery/decay phase weight modulation in ASP ensures that more prominent weights (corresponding to common features across different input patterns) are forgotten less. This in turn helps to retain the common features (or weight connections) from old data (forgetting connections that are specific to the old data) while learning a new input pattern making ASP learning more generic. Note, the values for all the hyperparameters used in Eqn. 6-8 are shown in Table I. 
\begin {table}[H]
\caption {Parameter Table} 
\begin{center}
\begin{tabular}{ |c|c| } 
 \hline
\textbf{Hyperparameters} & \textbf{Value} \\ 
\hline
$\tau_{rec}$ & 4 ms \\ 
\hline
$\tau_{acc}$ & 40 ms \\ 
\hline
$\tau_{post}$ & 80 ms \\ 
\hline
$offset$ & 0.2 \\ 
\hline
$k_{const}/k_{1,const}$ & 0.01 \\ 
\hline
$k_{2,const}$ & 1e2 \\ 
\hline
$\alpha, \alpha_{lin}$ & 0.01 \\ 
 \hline
\end{tabular}
\end{center}
\end{table}

\section{Experiments}
\subsection{Simulation Methodology}
The ASP learning algorithm was implemented in BRIAN \cite{goodman2008brian} that is an open source large-scale SNN simulator with parameterized functional models (LIF) for spiking neurons. We used the hierarchical SNN framework to perform digit recognition with the MNIST dataset \cite{lecun1998gradient}. The network topology and the associated synaptic connectivity configuration were programmed in the simulator. The spiking activity (or time instants of spikes) of pre- and post-neurons were monitored to track the corresponding pre/post synaptic traces that were used to estimate the weight updates in the recovery/decay learning phase of ASP.  
As mentioned earlier (in Section 4.3.1), we use the post-synaptic spike triggered weight update to speed up the simulation.

The effectiveness of the ASP algorithm was found to be dependent on the following system parameters that had to be tuned in a holistic manner for \textit{stable-plastic} learning.
\begin{itemize}
\item ASP recovery phase parameters, namely, the decay time constants of the accumulative/recent presynaptic trace and postsynaptic trace.
\item ASP decay phase parameters, namely, the decay rate of the weights ($\alpha$, $\alpha_{lin}$). The decay rate cannot be too high as it would result in faster weight leaking prohibiting the neurons from learning any representation. Very low decay rate will cause the connections to forget slowly that would result in subsequent overlapping of representations while learning new patterns. We used a nominal $\alpha$ value = 0.0001 for exponential decay and  $\alpha_{lin}$ value = 0.01 in our simulations.
\end{itemize}

Please note, we use identical parameters for neuron and synapse models, homeostasis, input encoding and input image presentation time as Diehl \& Cook \cite{diehl2015unsupervised} for fair comparison of our ASP learning with standard STDP learning. The standard STDP learning model is implemented using the power law weight dependent rule \cite{diehl2015unsupervised, querlioz2013immunity}. Furthermore, across all comparative experiments between standard STDP and ASP described below, the SNNs in both cases were trained using the same number of training images with equivalent dynamic environment setup. After the training is complete, each excitatory neuron of the SNN encodes generalized representation of the individual patterns as shown in Fig. 4 (a). During testing, the neurons that have learnt a particular digit fire steadily for a corresponding test input based on which the predicted class is inferred. This inference approach is similar to that of Diehl \& Cook.

\subsection{Learning to forget with ASP in a dynamic digit recognition environment}

\begin{figure}[t!]
\centering
\subfloat[]{\includegraphics[width=0.5\textwidth]{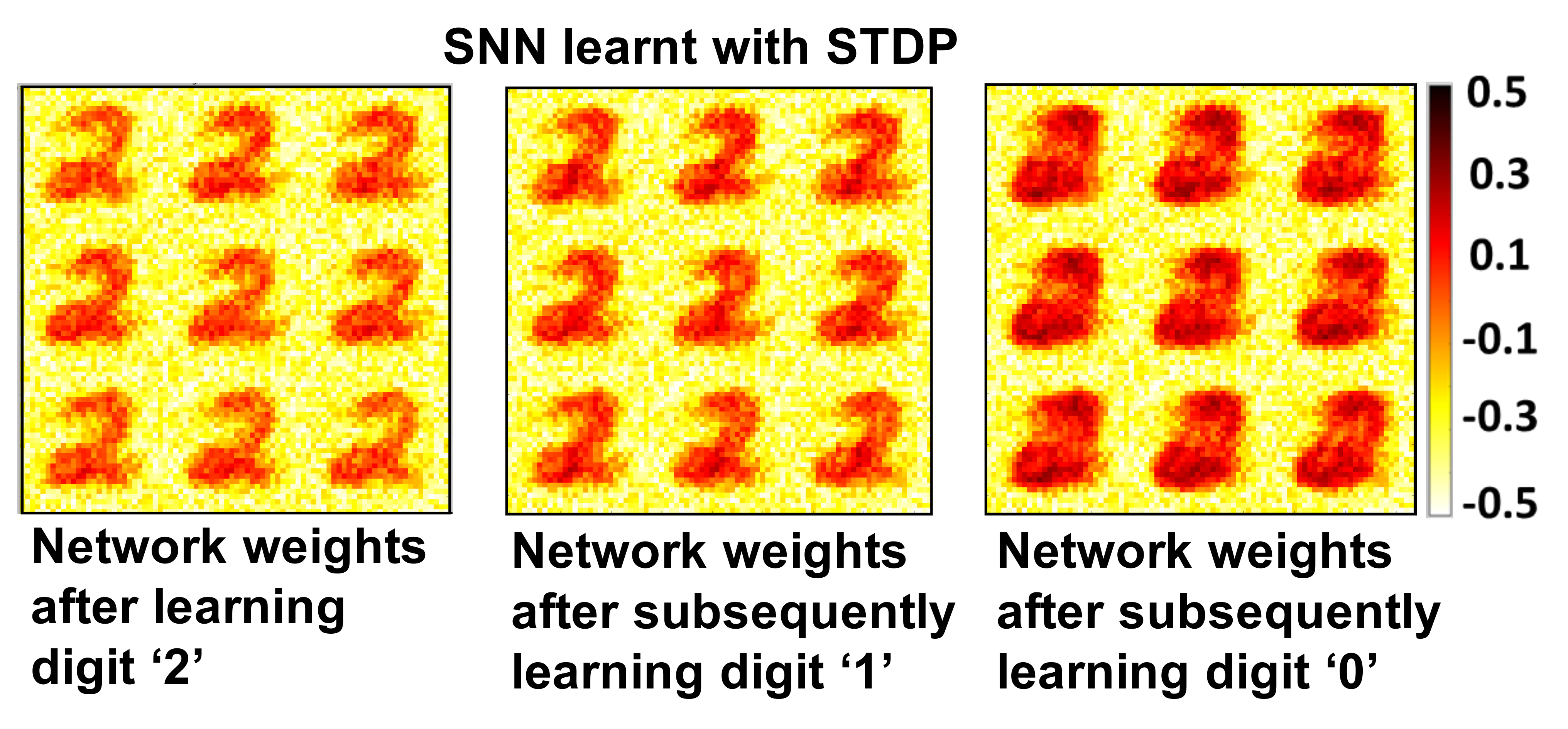}
\label{(a)}}
\hfill
\subfloat[]{\includegraphics[width = 0.5\textwidth]{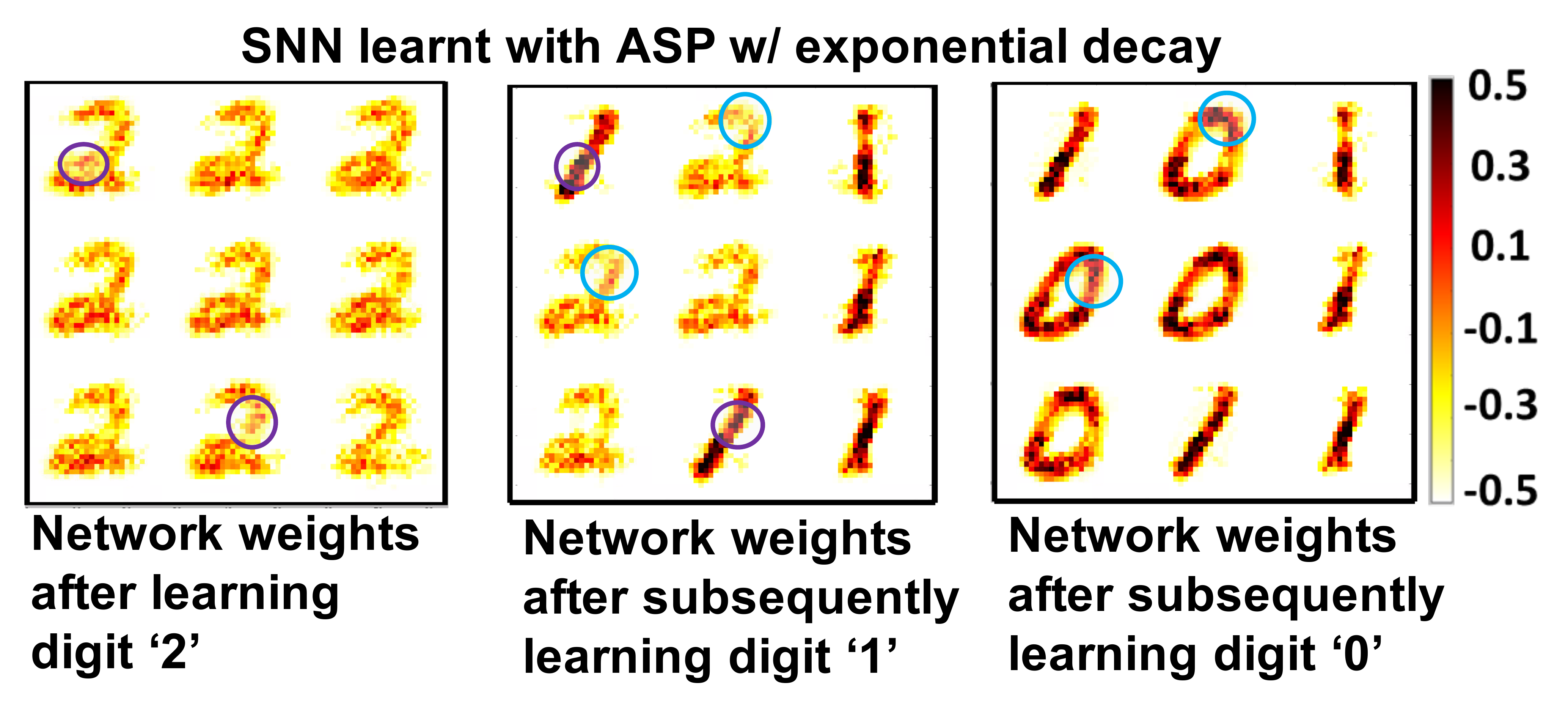}
\label{(b)}}
\hfill
\subfloat[]{\includegraphics[width = 0.5\textwidth]{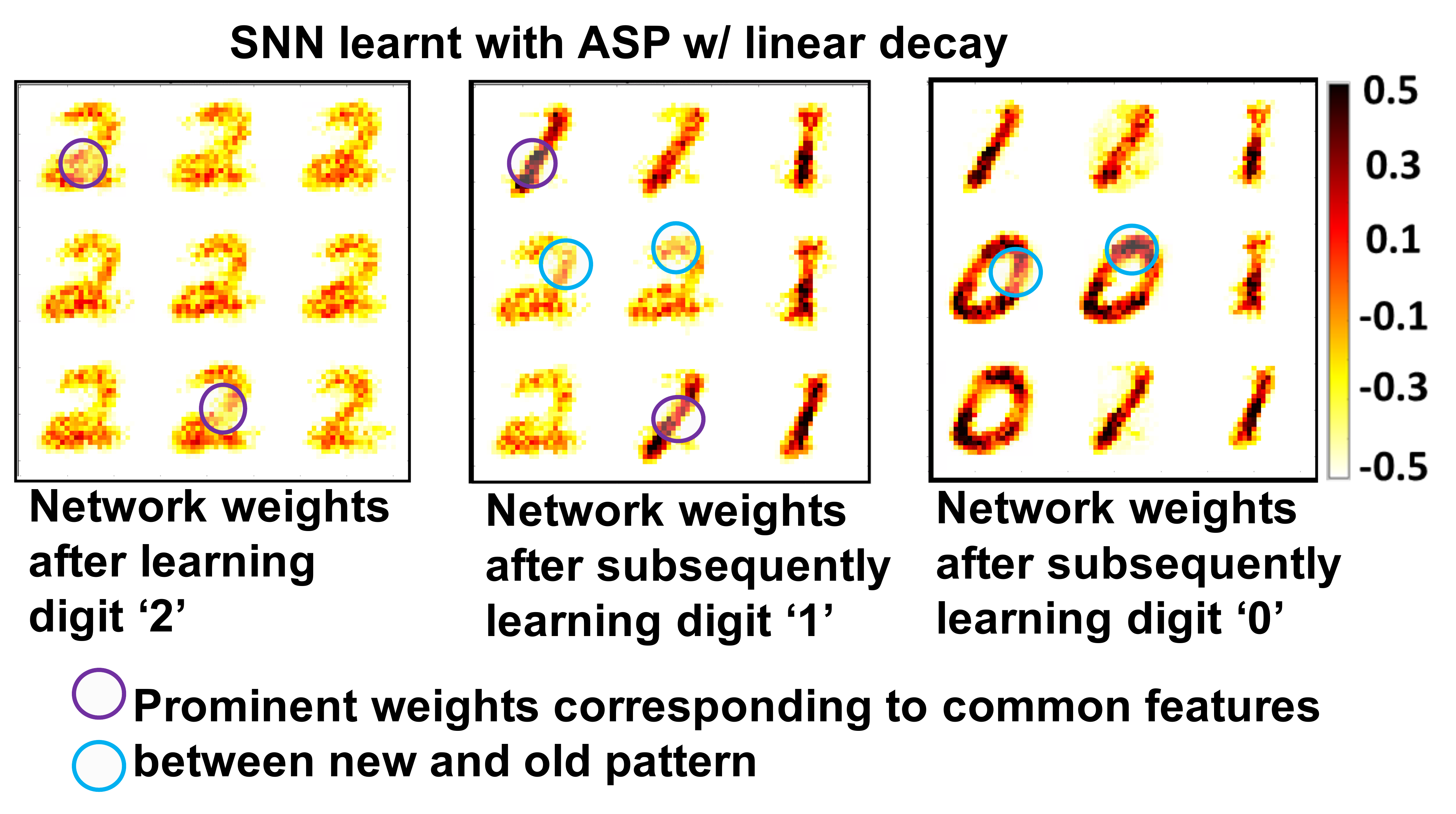}
\label{(c)}}
\caption{Digit representations learnt in a dynamic environment with digits '2'  through '0' shown sequentially to an SNN (with 9 excitatory neurons) (a) learnt with STDP (b) learnt with ASP with exponential decay (c) learnt with ASP with linear decay. Prominent weights corresponding to common features across different categories that are accentuated during the learning process with ASP have been selectively marked.}
\label{fig9}
\end{figure}

The main motivation for ASP is to come up with a biologically inspired learning paradigm that will facilitate on-line and adaptive learning in non-stationary environments. 
In order to create a dynamic environment with digit recognition framework, we presented the training instances of digits `0' through `9' sequentially with no reinforcement i.e. no training image is re-shown to the network. The digit categories are shown one-by-one with no intermixing of categories at any time during the training phase. Thus, the network must learn to forget old digit categories and retain the more recent  ones while trying to learn a new category. 

Fig. \ref{fig9} shows the representations learnt in a fixed-size SNN (with 9 excitatory neurons) with traditional STDP learning against our proposed ASP learning (both exponential and linear decay) in a dynamic environment (wherein digits `2' through `0' are presented sequentially in the order as `2$\rightarrow$1$\rightarrow$0'). We see that as the network is shown digit `1', ASP learnt SNN (Fig. \ref{fig9} (b), (c)) forgets the already learnt connections for `2' and learns the new input. Also, ASP enables the SNN to learn more stably as some neurons corresponding to the older pattern `2' are retained while learning `1'. When the last digit `0' is presented to the ASP learnt SNN, the connections to the excitatory neurons that have learnt digit `2' are forgotten to learn ‘0’ while the connections (or neurons) corresponding to recently learnt digit `1' remain intact. This is in coherence with our significance and latest data driven forgetting mechanism (incorporated in the decay phase in ASP) wherein older digits are forgotten to learn new digits. However, with STDP learnt SNN, the representations overlap thereby rendering the network useless towards the end of training. It is worth mentioning that the network representation after learning digit `1' in Fig. \ref{fig9} (a) is not very different from the prior representation after learning digit `2'. This indistinguishability arises as the network weights corresponding to `1'  overlap with the previously learnt weights of `2'. Specifically, the overlap occurs in the inclined region of `2' thereby making both the SNN configurations visibly similar. 

A notable observation in Fig. \ref{fig9} (b), (c) (ASP learnt SNN) is that a new digit (`1'/ `0') is learnt by forgetting specific features that have no similarity with the new pattern and retaining (and accentuating) common features from the old pattern `2'). As discussed earlier, the common features are encoded in the prominent weights (learnt during recovery phase of ASP). In Fig. \ref{fig9}, we can see from the color coding that the prominent weights (or common features between digit `2' and `1'; digit `2' and `0') have a higher intensity (almost black). This shows that ASP learns more generic representations associating relevant information (or connections in this context) from already learnt data to new data. Please note that similar behavior is observed in both exponential and linear decay based ASP. 

\begin{figure*}[t!]
\centering
\subfloat[Dynamic environment (no data reinforcement)]{\includegraphics[width=0.6\textwidth]{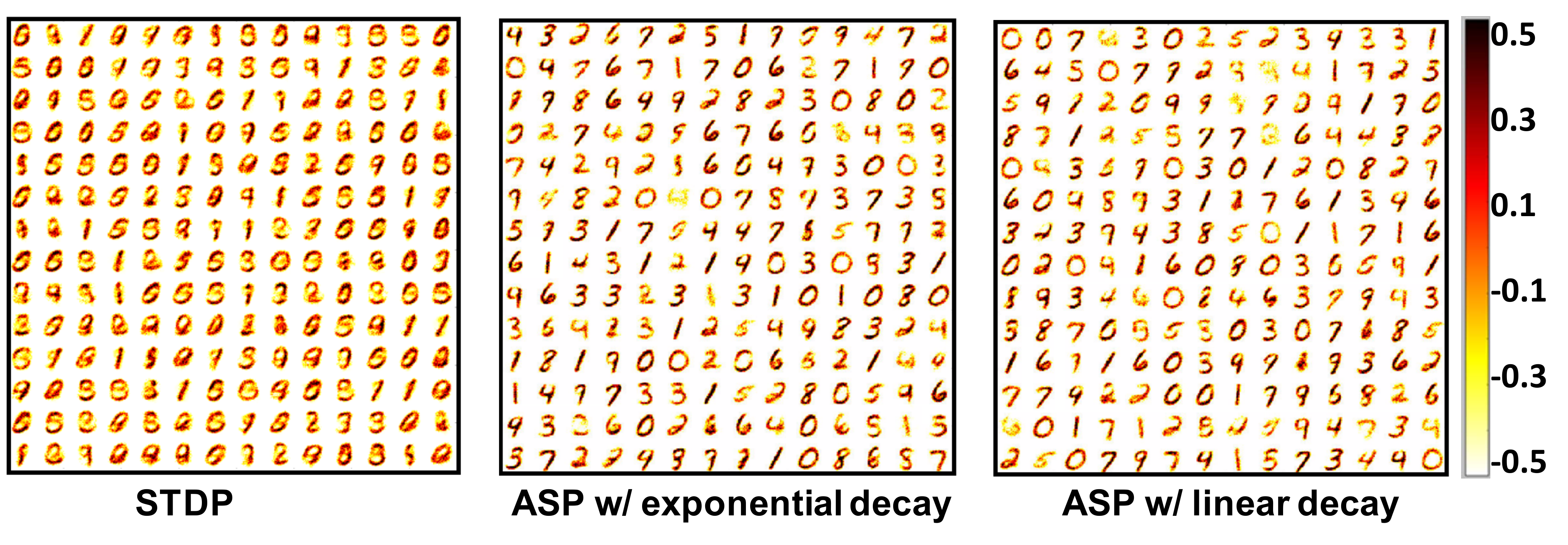}
\label{(a)}}
\subfloat[]{\includegraphics[width=0.4\textwidth]{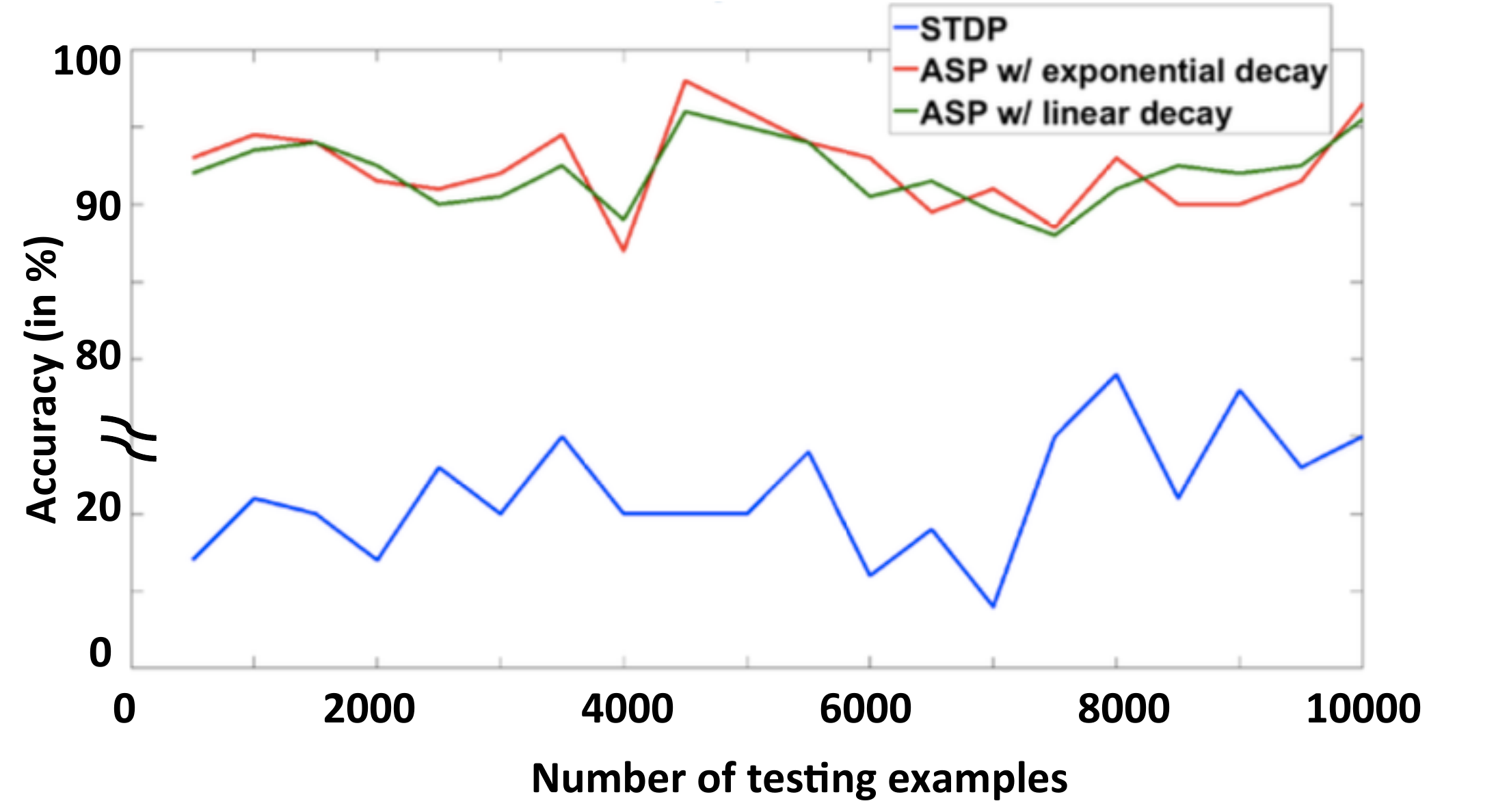}
\label{(b)}}
\caption{(a) Digit representations (shown for a sample of 196 neurons) learnt in a dynamic environment with digits '0'  through '9' shown sequentially to an SNN (with 6400 excitatory neurons) with ASP and traditional STDP learning. (b) Classification accuracy obtained from the networks trained on the different learning models in dynamic environment  as the number of test instances shown to the network are varied.}
\label{fig10}
\end{figure*}

\begin{figure*}[t!]
\centering
\subfloat[Data reinforcement]{\includegraphics[width=0.6\textwidth]{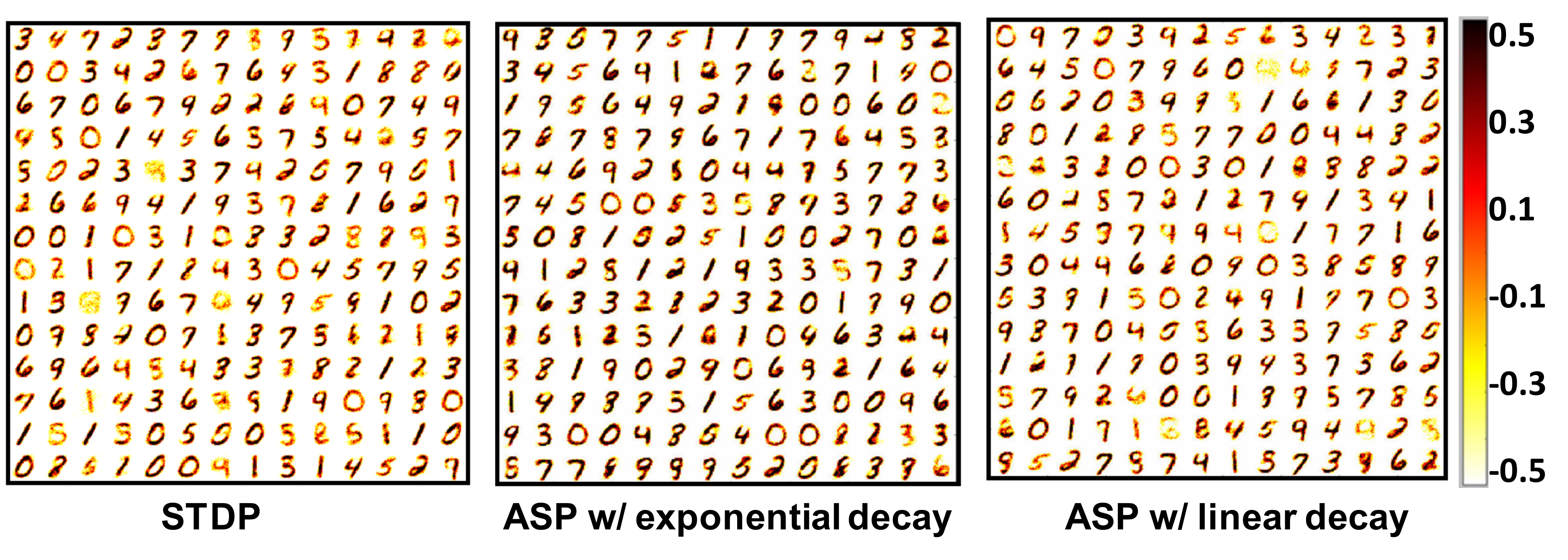}
\label{(a)}}
\subfloat[]{\includegraphics[width=0.4\textwidth]{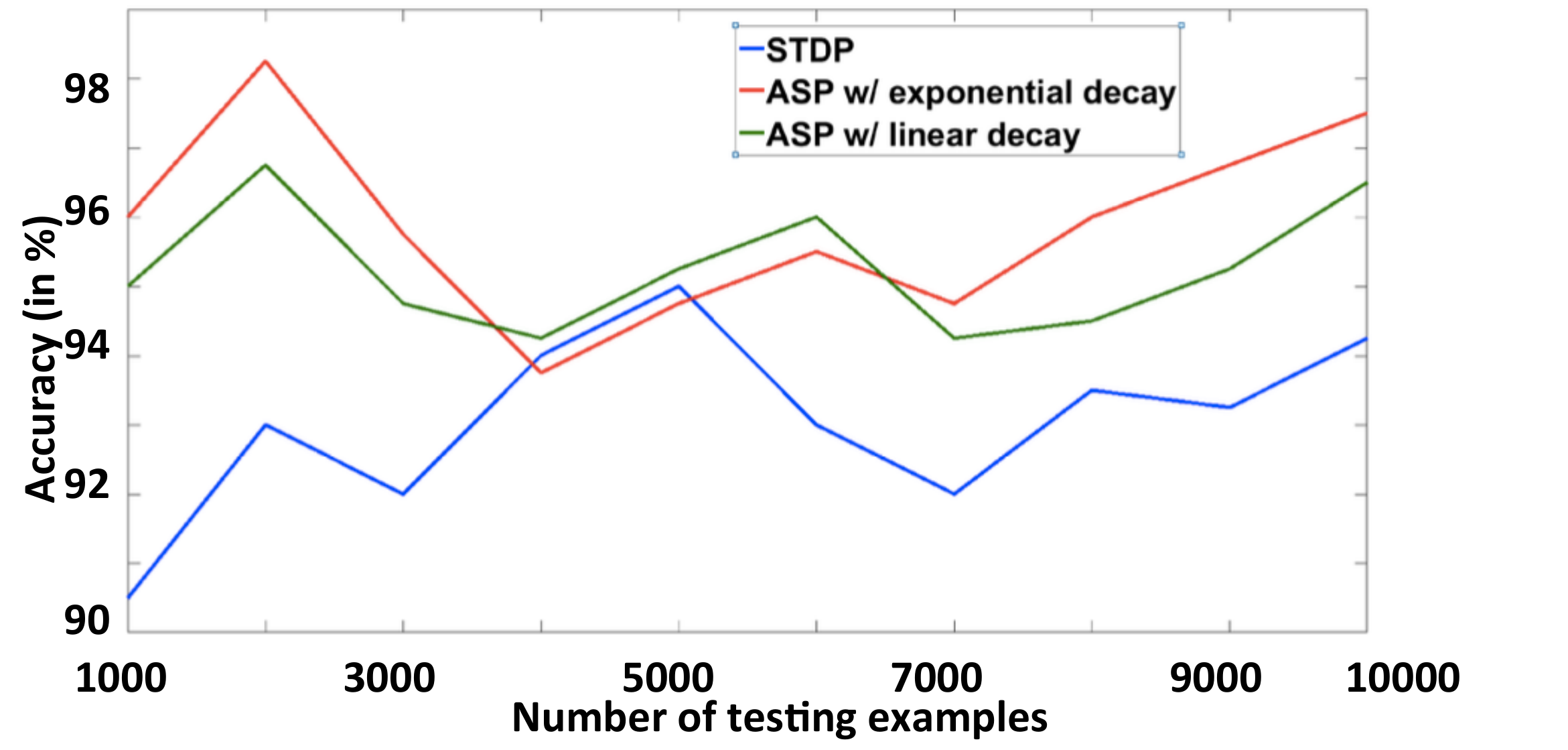}
\label{(b)}}
\caption{(a) Digit representations (shown for a sample of 196 neurons) learnt in a data reinforced environment with digits '0'  through '9' presented in an intermixed manner to an SNN (with 6400 excitatory neurons) with ASP and traditional STDP learning (b) Classification accuracy obtained from the networks as the number of test instances are varied in data reinforced environment.}
\label{fig11}
\end{figure*}

\begin{figure}[t!]
\centering
\includegraphics[scale =0.55]{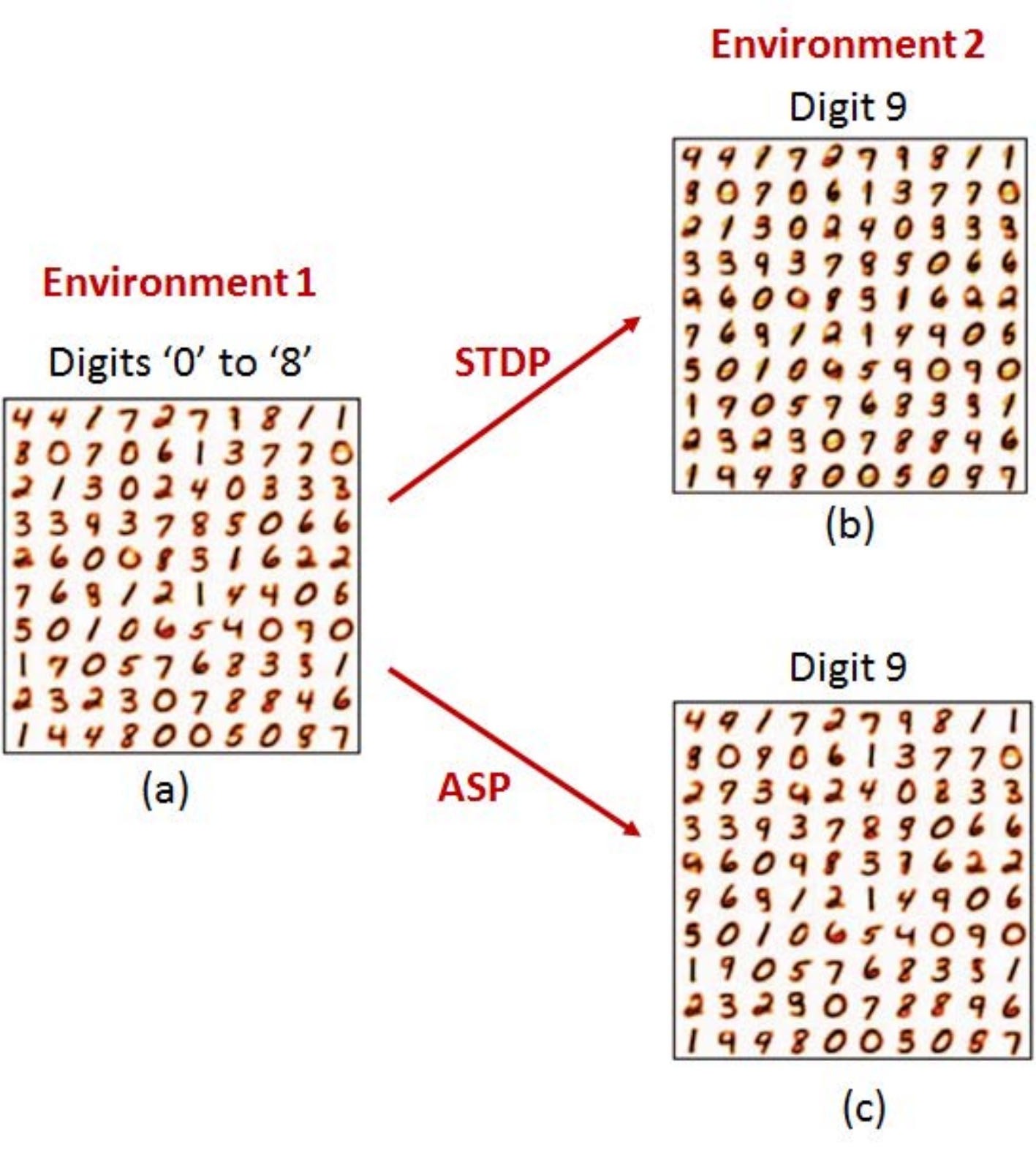}
\caption{(a) SNN after learning digits ‘0’ through ‘8’ with data reinforcement (b) The same SNN from (a) after being presented with digit ‘9’ learnt with traditional STDP(c) The same SNN from (a) after being presented with digit ‘9’ learnt with ASP. }
\label{fig11(c)}
\end{figure}

Fig. \ref{fig10} (a) shows SNNs (with 6400 excitatory neurons) learnt with ASP (both exponential and linear decay) and standard STDP in a dynamic environment when all digits `0' through `9' are presented sequentially. To ensure that the earlier digits are not completely forgotten, the number of training instances of each digit category were arranged in a decreasing order i.e. digit `0' had more training instances than digit `1' and so on. It has been stated earlier that the number of times a particular category is shown to the SNN also quantifies the significance of that digit. So, the network will try to retain more significant data while learning recent patterns. It is clearly seen that the SNN learnt with our proposed ASP encodes a better representation of the input patterns in comparison to the standard STDP trained network. In fact, the network is able to represent all digits suitably.  In the latter case, we observe that most of the representations are illegible due to substantial overlap. Fig. \ref{fig10} (b) shows the accuracy of the networks on the testing data from MNIST. While STDP yields a very low average accuracy of 23.30\%, ASP with exponential (or linear) decay has an accuracy of 94.85\% (or 94.2\%). This reckons the effectiveness of ASP (predominantly forgetting mechanism) for a dynamic on-line learning scenario. Furthermore, the comparable accuracy with exponential and linear decay ASP indicates that both types of decay can be effectively used to emulate the forgetting behavior and implement significance driven learning. 

\subsection{Accuracy improvement with ASP over standard STDP}
Till now, we have discussed how ASP enables an SNN to learn in a dynamic environment when the input digits are presented sequentially. 
As we saw earlier, SNN trained with standard STDP has significant overlap of representation when the inputs are presented without data reinforcement. 
For fair comparison for classification performance, we compare the accuracy for the two training methods when digits are presented to the SNN in an intermixed manner (i.e. with data reinforcement wherein the different digit categories are iteratively repeated over the training process). Fig. \ref{fig11} (a) shows the network weights at the end of training for a SNN with 6400 excitatory neurons learnt using ASP (both exponential and linear decay) and standard STDP. Fig. \ref{fig11} (b) illustrates the accuracy of the SNNs on the MNIST testing data. ASP learning yields an average test accuracy of 96.8\%/95.6\% for exponential/linear decay, respectively, that is $\sim$2.5\% higher than that of 94.3\% obtained with STDP learning. This proves that the forgetting behavior realized with ASP does not dominate the overall plasticity of synapses making the overall learning balanced. The improvement in accuracy is also indicative of ASP’s ability to generalize the network that causes it to learn more generic representations of the training data thereby avoiding overfitting. The slight difference in accuracy observed in the ASP case for linear and exponential decay can be attributed to the randomness in the Poisson based encoding of the input images.

To further elucidate the problem of sequential learning with STDP against ASP, let’s discuss two scenarios: In Fig. \ref{fig11(c)}, an SNN (with 100 excitatory neurons) is initially trained using STDP with data reinforcement (i.e. all image categories were intermixed) for the digits ‘0’ through ‘8’. In Scenario 1 continuing the STDP training, after this initial presentation of digits `0' to `8' (i.e. Environment 1), we subsequently presented the input image examples for the digit `9' without reinforcing or re-presenting any of the previous digits that resulted in data being overwritten. Fig. \ref{fig11(c)} (b) illustrates how weights associated with each neuron were affected in Environment 2 (i.e. only digit `9') with STDP learning, with many learned digits slowly transforming into the digit `9'. In Scenario 2, we learnt the same SNN from Fig. \ref{fig11(c)} (a) using ASP (instead of STDP) while showing only digit `9'. It is evident that changing the presentation of digits from Environment 1 to 2 is a dynamic or sequential learning scenario. Fig. \ref{fig11(c)} (c) shows the weights learnt in the SNN with ASP in Scenario 2. It is clearly seen that the network learns the digits `9' without any overlap while forgetting some old instances that are eventually transformed into digit ‘9’. Evidently, the SNN learnt with STDP (62.8\%) yields lower accuracy than that of ASP (73.4\%). The lower accuracy with ASP in this case as compared to Fig. \ref{fig11}/\ref{fig10} is due to the lesser number of excitatory neurons used in the SNN. Thus, in the absence of data reinforcement or retraining, training with STDP alone does result in catastrophic forgetting of old data in a sequential learning environment.

\subsection{Denoising with ASP}
\begin{figure}[t!]
\centering
\subfloat[AWGN]{\includegraphics[width=0.2\textwidth]{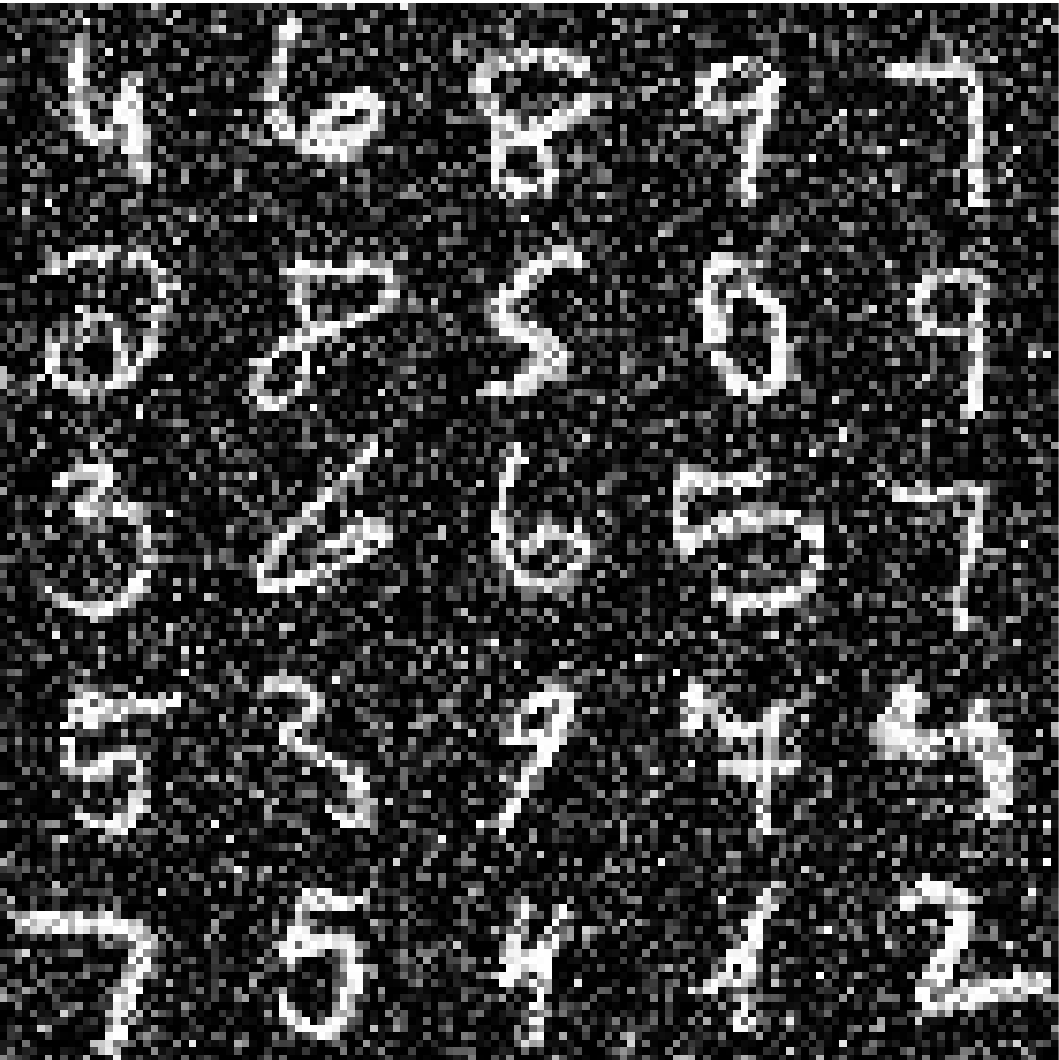}
\label{(a)}}
\hspace{2mm}
\subfloat[AWGN and reduced contrast]{\includegraphics[width=0.2\textwidth]{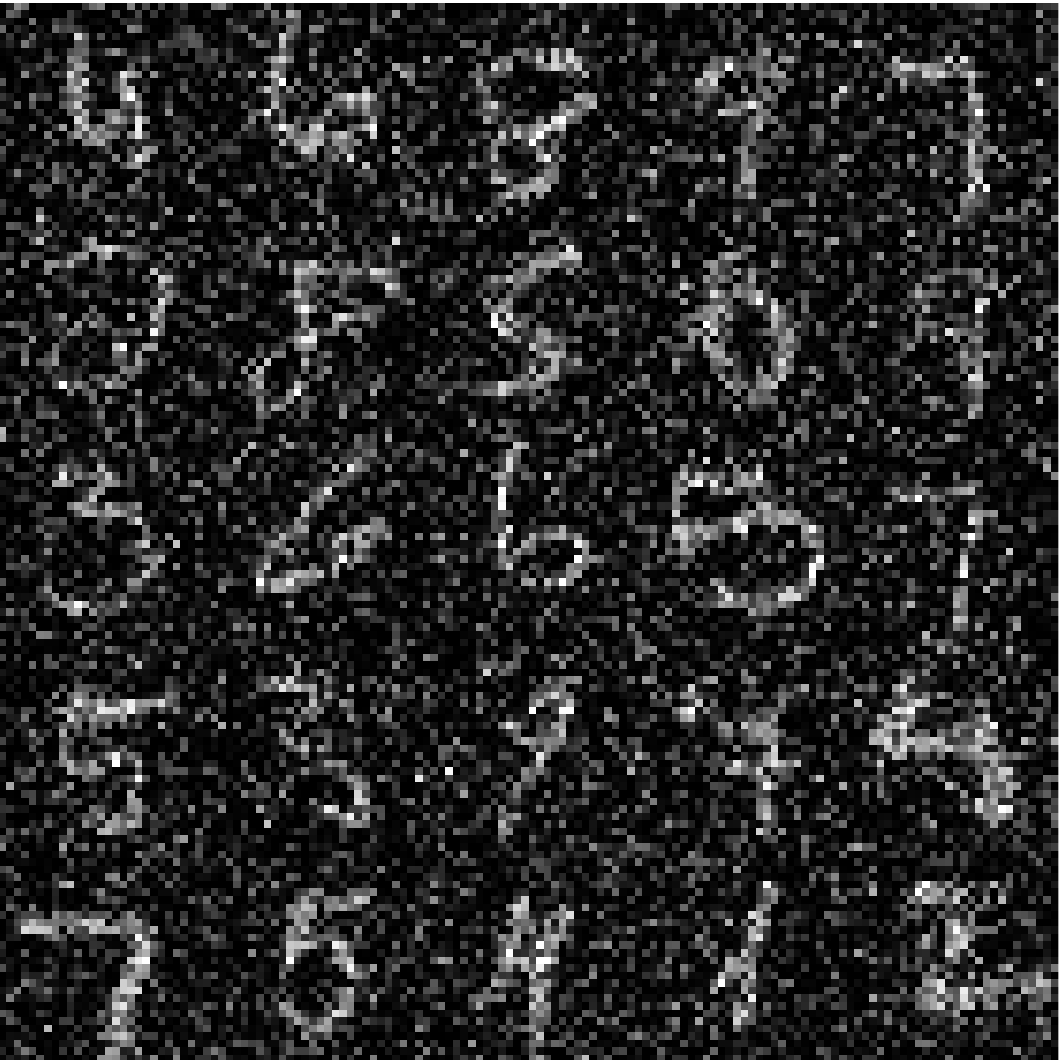}
\label{(b)}}
\caption{Noisy-MNIST images with (a) Additive White Gaussian Noise (AWGN) (b) Reduced Contrast with AWGN.}
\label{fig12}
\end{figure}

\begin{figure}[t!]
\centering
\includegraphics[width=0.5\textwidth]{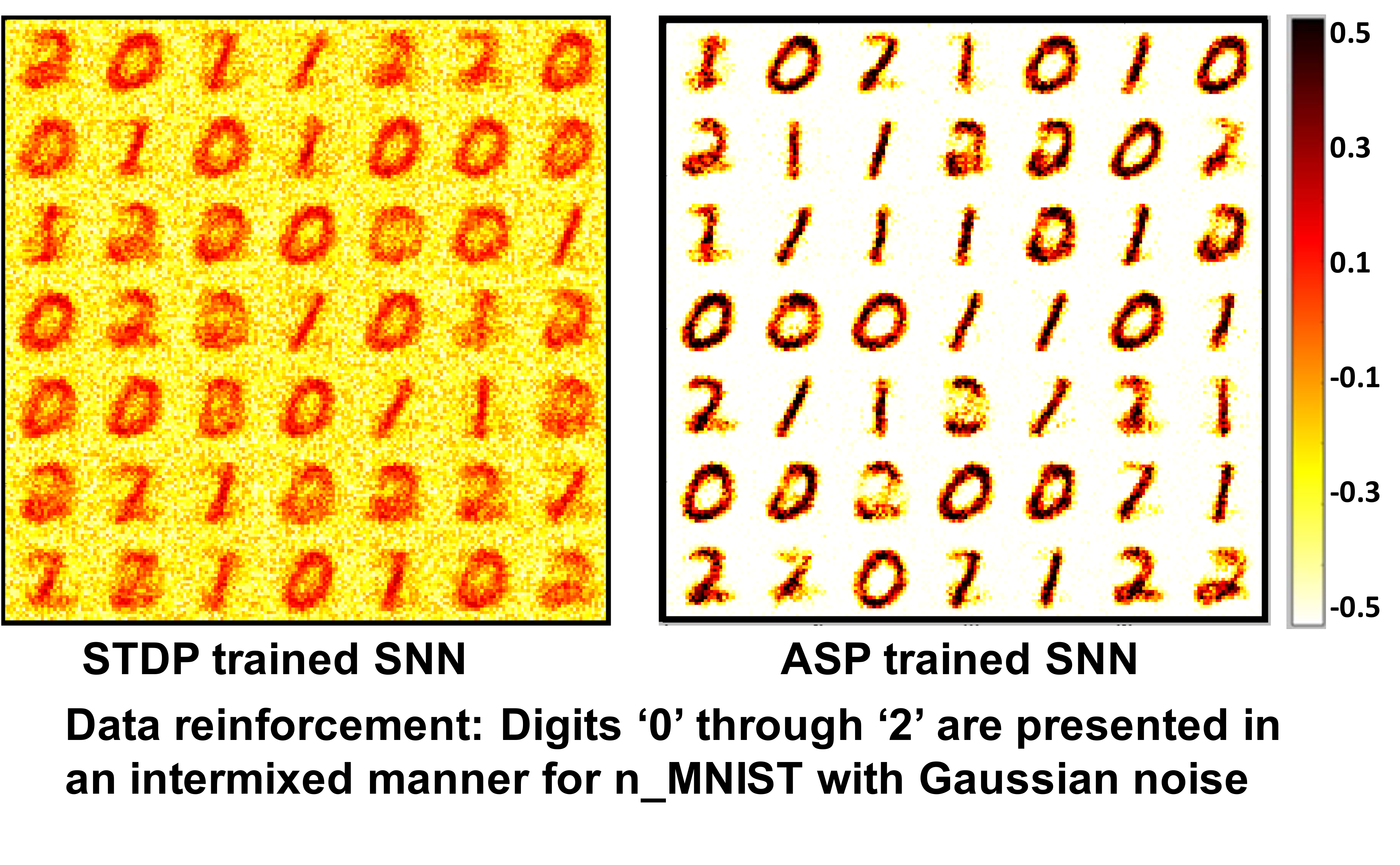}
\caption{Digit representations learnt with digits '0'  through '2' presented in an intermixed manner to an SNN (with 49 excitatory neurons) with STDP and ASP for Noisy-MNIST images with AWGN as Fig. \ref{fig12} (a).}
\label{fig13}
\end{figure}

\begin{figure}[t!]
\centering
\includegraphics[width=0.5\textwidth]{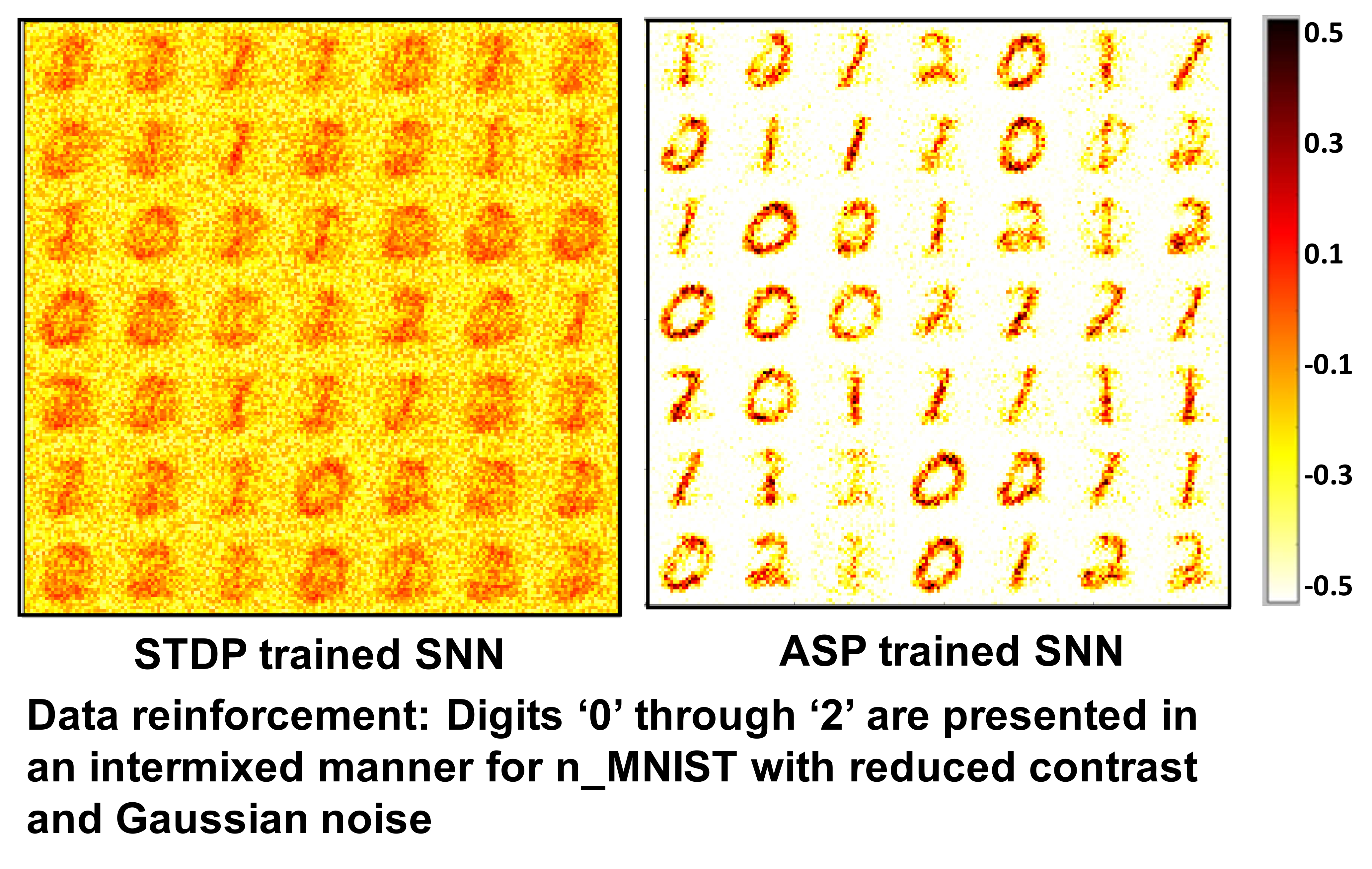}
\caption{Digit representations learnt with digits '0'  through '2' presented in an intermixed manner to an SNN (with 49 excitatory neurons) with STDP and ASP for Noisy-MNIST images with AWGN and reduced contrast as Fig. \ref{fig12} (b).}
\label{fig14}
\end{figure}

The decay incorporated in ASP besides rendering an SNN to self-adapt in a dynamic environment also offers a key advantage of denoising by extracting relevant features from noisy input images. To demonstrate this, we trained an SNN with Noisy-MNIST (n-MNIST) \cite{basu2015learning} images as shown in Fig. \ref{fig12}. Fig. \ref{fig12} illustrates the sample images from the n-MNIST dataset with (a) Additive White Gaussian Noise (AWGN) (b) a combination of AWGN and reduced contrast. Fig. \ref{fig13} shows SNNs (with 49 excitatory neurons) learnt with ASP and standard STDP when digits `0' through `2' are presented in an intermixed manner (with data reinforcement) for the n-MNIST images from Fig. \ref{fig12} (a). It is clearly seen that both STDP and ASP are able to encode the corresponding digit representations. However, ASP owing to the dynamic decay mechanism is able to complete eliminate the background noise while selectively attending (or learning) the task relevant information in the data. It is worth mentioning that, in the denoising experiments below, we compare the STDP trained SNN against ASP with exponential decay and show the results for the same. However, similar results can also be obtained for ASP with linear decay.

To further establish the adeptness of our ASP rule in reducing the influence of noisy inputs, we train an SNN on the n-MNIST images from Fig. \ref{fig12} (b) that have AWGN with reduced contrast. The intensity of the digit pixels are now equalized with that of the background noise pixels on account of the reduced contrast (as seen in Fig. \ref{fig12} (b)). Fig. \ref{fig14} shows the representations learnt with ASP and standard STDP for the same data-reinforced recognition scenario as that of Fig. \ref{fig13}. Now, we can clearly see that that, in spite of the reduced contrast, the ASP learning is able to distinguish the relevant information (or digit patterns) from the noisy background and learn robust representations. The weight decay phenomenon with varying leak rate in ASP  (refer to Section 4.3.2) retains significant information while forgetting (or leaking) the weights corresponding to irrelevant information thereby enabling the selective exclusion of the background noise. In contrast, standard STDP owing to the absence of dynamic weight decay is unable to filter out the relevant digit information from the background noise of similar intensity. Thus, the representation learnt in the latter case are mostly cluttered and obscure.  We would like to note that, in both cases of standard STDP and ASP, the SNNs were trained using the same number of training images. 

To quantify the classification performance on the n-MNIST images, we trained a larger SNN (with 6400 excitatory neurons) with standard STDP and ASP in a data-reinforced environment for digits `0' through `9'. STDP in both cases yields a lower accuracy (on the testing data from n-MNIST) than ASP. While the average accuracy with STDP is 87.4\% on just AWGN based noisy images (Fig. \ref{fig12} (a)), the accuracy degrades to ~52.1\% for the AWGN with reduced contrast images (Fig. \ref{fig12} (b)). ASP, on the other hand, consistently performs better with 93.8\% (85.6\%) classification accuracy for AWGN (reduced contrast with AWGN) n-MNIST images. This illustrates how the adaptive decay process incorporated in ASP automatically encodes selective filtering and attention towards task relevant features in the input data further improving the robustness of the unsupervised learning paradigm.


\section{Discussion \& Conclusion}
In this work, we proposed a novel bio-inspired unsupervised learning rule Adaptive Synaptic Plasticity (ASP) for real time on-line learning with Spiking Neural Networks (SNNs). We integrate weight decay with traditional STDP to devise our dynamic weight update ASP mechanism that addresses \textit{``catastrophic forgetting''}, a key issue across conventional learning models. The weight decay employed in ASP emulates the ``forgetting'' behavior of the mammalian brain. While STDP helps in learning new input patterns, the retention of significant, yet old, data or gradual forgetting of insignificant information is attained with weight decay. In ASP, we modulate the leak rate of the synaptic weight decay process using the temporal dynamics of pre- and post-synaptic neurons that maintains a balance between continuous learning and forgetting to construct a \textit{stable-plastic} self-adaptive SNN for evolving environments. ASP owing to its significance driven ``forgetting while learning'' formulation, enables an SNN to generalize over the training data with more generic learning of representations (thus avoiding overfitting) yielding significantly improved accuracy over traditional STDP methods. 

Further, we would like to point out that in order to prevent ``catastrophic forgetting'' with STDP learning in SNNs, the network is re-presented with the already learnt old information along with the new data, when the network has to learn a new class. However, storing all old data samples for retraining is a major drawback for implementing on-line real time learning. ASP offers a promising solution for real time dynamic learning without this retraining procedure. In addition to the dynamic learning, the adaptive weight decay mechanism in ASP also enables unsupervised denoising or selective attention towards relevant features in the input data further improving the robustness of the proposed learning.

Finally, the recent resurgence of neural networks after the \textit{artificial intelligence winter era} can be attributed to the learning of good, flexible representations from the visual world. In the current digital era, incorporating active sensing and diverse task modelling (classification, decision making, analytics) within the same learning model will no doubt shape the nature of learning representations in future cognitive systems. To that effect, the \textit{``learning to forget''} behavior realized with ASP (inspired deeply from biological principles) provides a potentially exciting and promising direction towards improved representation learning with the emerging computing paradigm of SNNs. 

\section*{Acknowledgment}
P.P., J.A. and K.R. are supported in part by C-SPIN, one of the six centers of StarNet, a Semiconductor Research Corporation Program, sponsored by MARCO and DARPA, by the Semiconductor Research Corporation, the National Science Foundation, Intel Corporation and by the DoD Vannevar Bush Fellowship. S.R. acknowledges ARO W911NF-16-1-0289 for support.

\end{document}